# The LLENS Platform for On-Premises Large Language Model Agents in a Japanese University Hospital: Development and Evaluation Study

Ken Enda[1,2,3], Jien Saito[4], Masami Mukai[1], Feng Han[1], Emir Turkes[1,5], Tatsumi Kusakabe[6], Reona Umeki[6], Kazuhiro Sumiyoshi[6], Mitsuya Numata[6], Takumi Tanikawa[6], Akira Endoh[6], Kenji Hirata[5,7], Atsushi Yoshida[1,5], Kohsuke Kudo[1,5], Shinya Tanaka[1,2,8]

1. Division of AI Support for Medical Research, Faculty of Medicine, Hokkaido University, Sapporo, Japan
2. Department of Cancer Pathology, Faculty of Medicine, Hokkaido University, Sapporo, Japan
3. Technoplasm Inc., Sapporo, Japan
4. Department of Cardiovascular Surgery, Graduate School of Medicine, Hokkaido University, Sapporo, Japan
5. Department of Diagnostic Imaging, Faculty of Medicine, Hokkaido University, Sapporo, Japan
6. Division of Medical Information Planning, Hokkaido University Hospital, Sapporo, Japan
7. Department of Nuclear Medicine, Faculty of Medicine, Kanazawa University, Kanazawa, Japan
8. Institute for Chemical Reaction Design and Discovery (WPI-ICReDD), Hokkaido University, Sapporo, Japan

Corresponding author: Ken Enda, Division of AI Support for Medical Research, Faculty of Medicine, Hokkaido University, Kita 15, Nishi 7, Kita-ku, Sapporo 060-8638, Japan. Phone: +81-11-706-5352. Email: kenda@med.hokudai.ac.jp

ORCID iDs: Ken Enda, 0000-0002-4398-3292; Jien Saito, 0000-0003-2484-258X; Masami Mukai, 0000-0001-6315-6078; Feng Han, 0000-0002-6664-0516; Emir Turkes, 0000-0002-7447-1476; Takumi Tanikawa, 0000-0002-5067-4818; Kenji Hirata, 0000-0003-0036-8975; Atsushi Yoshida, 0000-0001-5336-9985; Kohsuke Kudo, 0000-0001-5351-9242; Shinya Tanaka, 0000-0001-6470-3301.

## Abstract

**Background:** Large language model agents that use tools could take on part of the electronic health record and desk work that occupies much of physicians' time. In many hospitals, however, medical information systems, including the electronic health record, are on-premises and run inside closed networks, so the model and the agent harness, which reads materials, computes, and returns artifacts, must also be inside the network.

**Objective:** We designed, implemented, and deployed LLENS (Large Language Enhanced Nexus System), a clinical support artificial intelligence agent platform that runs inside the closed network of Hokkaido University Hospital, and report its development and technical evaluation: model selection and serving, question answering over hospital documents, and the harness components and scenarios built from them.

**Methods:** LLENS runs on a single host with 8 NVIDIA H200 graphics processing units, the open-weight model Kimi K2.6 (~1 trillion parameters), and a web application (Nucllei) rebuilt from Open WebUI. The agent harness has 10 components, including page image reading, full-text attachment input, a path-addressed document tree without a vector index, a network-isolated sandbox, artifact return, self-checking of generated output, subagents, and saved procedures. The model was chosen from 3 candidates by Japanese medical benchmark scores, weighing image input, context length, and key-value cache capacity for concurrency, and serving performance was measured at 4-16 concurrent requests. For hospital-document access, the model answered 180 Japanese questions written by the development team (3 seeds each) with no documents, fixed retrieval (top 10 chunks of one lexical search using the question), or interactive access (the model reading with document tools and sandbox); grading was automatic. Component triggering was tested on 13 work scenarios (4 runs each).

**Results:** Kimi K2.6 scored 480/500 on the 2026 Japanese Medical Licensing Examination. For a single user, it generated 119 tokens/s with thinking and 129 tokens/s without, and served 16 concurrent requests at 908 tokens/s in total. Mean accuracy was 10.7% with no documents, 81.3% with fixed retrieval, and 97.2% with interactive access, 15.9 percentage points above fixed retrieval (95% CI 10.9-21.3; *P*<.001). The difference lay in filtering/counting questions (fixed retrieval 16.7% vs interactive access 87.8%). The expected components were triggered in all 4 runs for 12 of the 13 scenarios, and 5 scenarios were run with synthetic data on the deployment host. The platform has run on the hospital network since September 14, 2026.

**Conclusions:** With a trillion-parameter-class open-weight model and the harness, an agent on a single host inside a closed hospital network carried out multistep work in which it searched hospital documents itself, computed, and returned artifacts. Future work will connect the electronic health record and departmental systems so that the model reads patient records directly instead of exported files.

# Introduction

A systematic review of studies in inpatient and outpatient settings found that physicians spend on average about 37% of their working time on the electronic health record [1], and an observational time study of physicians in Japanese hospitals found that writing medical records and other documents and entering orders took about one-fifth of their clinical working time [2]. If routine work around clinical judgment, such as writing summaries and registering data, could be handed to artificial intelligence (AI), physicians would have more time for patient care and research. Large language models (LLMs) that use tools are becoming able to carry out such work themselves as agents [3,4].

Doing this in a hospital, however, is subject to a major constraint: the handling of patient information [5]. In Japan, medical information systems, including the electronic health record, are mostly on-premises [6] and run inside closed medical information networks [7], and Hokkaido University Hospital is no exception. Processing of patient-derived data stays inside this closed network, so external cloud application programming interfaces (APIs) cannot be used. The mechanism that reads hospital materials and users' files, computes, and returns artifacts has to sit inside the closed network together with the model.

Hospital work also goes beyond "search documents and summarize them," as in looking up hospital rules. It includes, for example, reading scanned forms, aggregating a list of laboratory values into a chart, preparing Word or Excel templates, reading and comparing several materials, and saving a procedure that worked so that it can be used next time. A chat screen and a search function alone cannot do this work. The work needs a mechanism that determines what goes into the model, which tools it is allowed to use, and what it is made to return. We call the set of components that make up this mechanism the agent harness. Each component is an existing technique, but in combination they can carry out work scenarios.

Most reports of running a model inside a hospital use a small model for a specific purpose [5,8,9]. Agents deployed in hospitals are also fixed to one use, such as drafting documents within the electronic health record [10,11]. A general-purpose in-hospital LLM for all staff has been reported [12], but we found no report of an agent with components that read materials, compute, and return artifacts, run inside a closed network as a general-purpose platform.

We designed and implemented LLENS (Large Language Enhanced Nexus System), a clinical support AI agent platform that carries a trillion-parameter-class open-weight model and the harness on a single host inside the closed network, and deployed it in Hokkaido University Hospital. LLENS was developed at the Faculty of Medicine, Hokkaido University, as the core of PRISM-HU (Strategic Promotion of Next-Generation AI-Enabled, Data-Driven, and Information-Circulating Medical Research through the Integrated Development of People, AI/DX [digital transformation], and Systems), a Hokkaido University project selected for a program of the Japan Agency for Medical Research and Development (AMED) in 2025 [13,14]. This paper shows that a general-purpose agent platform can be built and operated inside a closed network, and reports the design and deployment of the platform, model selection and serving performance, question answering over hospital documents compared with fixed retrieval, the correspondence between harness components and work scenarios, and 5 demonstrations on the deployment host. The connection to the electronic health record is being built, and the effects on users and on work will be evaluated alongside it (Discussion, "Next Steps").

# Methods

## Model and Serving

The system prompt, skills, and tool descriptions of the harness are written in natural language, and how they are read and followed depends on the ability of the model. Some components, such as image input, depend on a feature of the model itself. For these reasons, the model was selected first. The host is a single server with 8 NVIDIA H200 NVL graphics processing units (GPUs; 141 GB each, 1,128 GB in total; see "Architecture"), and the candidates were open-weight models with native function calling that run in this capacity.

Three open-weight candidates, Kimi K2.6, GLM-5.1, and DeepSeek V3.2 [15–17], were compared on their scores on Japanese medical benchmarks, image input, context length, the key-value (KV) cache capacity that can be given to concurrent requests, and license. Table 1 shows the specifications of the 3 candidate models. The main benchmarks were the 120th (February 2026) and 119th (2025) Japanese Medical Licensing Examinations (JMLE2026 and IgakuQA119 [18,19], each with 400 questions weighted to a maximum of 500 points). The 120th examination was held after the release of DeepSeek V3.2 (December 2025) and before the releases of Kimi K2.6 and GLM-5.1 (April 2026). We used the prompts and answer extraction of the benchmark providers (for the 2025 examination, we corrected answer-extraction errors in 2 questions and rescored 1 question for which the Ministry of Health, Labour and Welfare accepted 2 choices) and report the weighted scores. The candidates were run with and without thinking (a mode in which the model generates its reasoning before the answer), at temperature 0, with an output limit of 32,768 tokens and 1 response per question. Kimi K2.6 was given the question images, and the others were given text only. To compare the candidates with matched inputs, we also computed the scores on the text-only questions (Table S1 in Multimedia Appendix 1). As secondary benchmarks, we also measured the 3 tasks of JMED-LLM [20], the 4 tasks of llm-jp-eval [21], and IgakuQA 2018-2022 [22] (public for years and possibly in the training data), with and without thinking.

The deployment configuration adds EAGLE-3 to Kimi K2.6 (speculative decoding in which a small draft model proposes the next tokens and the main model verifies them together [23,24]), with a maximum context of 262,144 tokens and up to 16 concurrent requests. In this configuration, the key-value cache holds 695,195 tokens. The serving tests used SGLang 0.5.17. On the same host and both with and without the draft model, we sent synthetic requests of up to 1,024 input and 512 output tokens (lengths drawn uniformly up to these limits), 32, 64, and 128 requests at 4, 8, and 16 concurrent requests, respectively. We also recorded the single-user decode speed (3 short Japanese prompts with 600 output tokens each), an input-length sweep with a single request (without thinking, under conditions in which the prefix cache does not take effect), and the GPU power.

Table 1. Specifications of the candidate models. All 3 are MoE models. The total parameters were counted from the distributed weights and include the MTP layers of GLM-5.1 and DeepSeek V3.2 (10.0 B and 13.5 B). The active parameters were also counted from the distributed weights. The weight size is the total size of the distributed weight files. Release dates are from each developer's model repository. FP8: 8-bit floating point; INT4: 4-bit integer; MoE: mixture of experts; MTP: multi-token prediction.

| | Kimi K2.6 (selected) | GLM-5.1 | DeepSeek V3.2 |
|---|---|---|---|
| Developer | Moonshot AI | Z.ai (Zhipu AI) | DeepSeek |
| Total / active parameters | 1,027 B / 32 B | 754 B / 41 B | 685 B / 37 B |
| Weight precision (as distributed) | INT4 | FP8 | FP8 |

| | Kimi K2.6 (selected) | GLM-5.1 | DeepSeek V3.2 |
|---|---|---|---|
| Weight size (distributed files) | 595 GB | 756 GB | 689 GB |
| Maximum context (model) | 262,144 | 202,752 | 163,840 |
| Context setting in the benchmark runs | 131,072 (thinking on) / 262,144 (thinking off) | 131,072 | 163,840 |
| Speculative decoding | EAGLE-3 (separate draft model) | MTP (included in the weights) | MTP (included in the weights) |
| Image input (native) | Yes | No | No |
| License | Modified MIT | MIT | MIT |
| Public release | April 2026 | April 2026 | December 2025 |

## Architecture

Figure 1 shows the architecture of the platform. In addition to the 8 GPUs described in “Model and Serving,” the host has 2 AMD EPYC 9355 processors and 1.5 TiB of memory, and it is placed inside the hospital’s closed medical information network. Inference is run by SGLang [25] with tensor parallelism across the 8 GPUs. The web application Nucllei is a fork of Open WebUI v0.6.5 [26], developed jointly with Technoplasm Inc. and published as source under terms based on the PolyForm Noncommercial License 1.0.0 [27]. From the base, we removed the built-in retrieval-augmented generation (embeddings and vector store), the knowledge base, the built-in embedding-based memory across conversations, web search, image generation, in-browser code execution, prompt management, and sharing with the external community, and we added skills, built-in tools, subagents, and asynchronous execution. Tools are not chosen by the user for each chat but fixed on the server side for each model preset (a set of model, system prompt, and tools). Hospital documents are provided by a document service that runs as a separate process, and code execution is handled by an isolated sandbox.

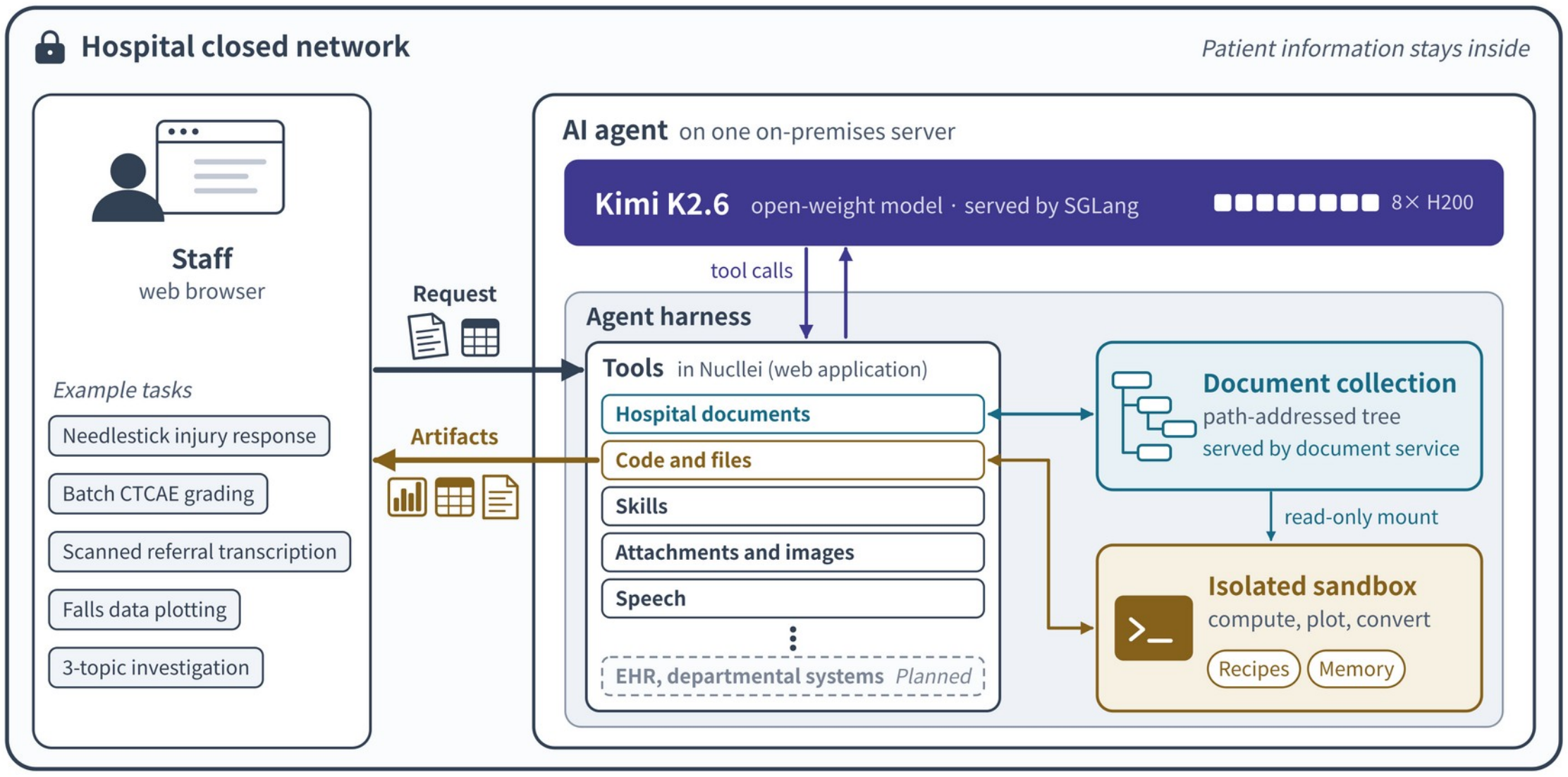

*Figure 1. Architecture of the LLENS platform. On one server inside the closed network, the model runs with an agent harness made of the web application Nucllei, the document collection, and the sandbox. Nucllei receives the staff member's request, calls the model, executes the tool calls the model returns, and returns the artifacts. Table 2 lists the tools, and the example tasks are the 5 scenarios demonstrated on the deployment host. AI: artificial intelligence; CTCAE: Common Terminology Criteria for Adverse Events; EHR: electronic health record.*

## Harness Components

The harness is the set of components that includes the system prompt, tools, and skills given to the model and the document service, sandbox, and other services behind the tools. The system prompt sets the policy for responding to users (mainly physicians), the scope of what is handled, how evidence is to be shown, and how the components below are to be used. Table 2 summarizes the tools the model can call.

The components were derived by decomposing the work scenarios we envisioned. There are 13 scenarios: the 5 demonstrated on the deployment host (an inquiry about the initial response to a needlestick injury, batch grading by the Common Terminology Criteria for Adverse Events (CTCAE), scanned referral letter transcription, falls data aggregation and plotting, and investigation of 3 topics by subagents) and 8 envisioned scenarios that can be built from the same components. The envisioned scenarios are the analysis of research data (a baseline characteristics table, survival analysis, and figures), transcription and trend charts from scanned laboratory report sheets, the conversion of legacy Word, Excel, Access, and compressed files, the calculation and checking of clinical scales from their defining formulas, the checking of medical fee billing requirements, the summarization of long materials, minutes from meeting recordings, and the sharing and rerunning of procedures. The components fall into 10 kinds: (1) reading images and Portable Document Format (PDF) files without a text layer as page images; (2) putting the full text of an attachment into the context, or expanding it into the sandbox when it is too large to fit; (3) a document tree addressed by path; (4) an isolated execution environment (sandbox) with bundled packages; (5) returning the artifacts the model has made to the user as files; (6) the model rechecking the figures and documents it has made by reading them as images (self-checking of generated output); (7) having a subagent read materials and receiving only its conclusion; (8) guides provided by the developers (skills); (9) procedures that users save (recipes); and (10) memory about the user. The correspondence between scenarios and components is shown in Results, “Scenarios and Components”.

Table 2. Tools given to the model (deployment configuration at the time of submission). Administrator-only operational tools (monitoring) are excluded.

| Category | Tool | What it does |
|---|---|---|
| Attachments and images | `read_attachment` | Reads the text extracted from an attachment, one range at a time |
| | `feed_image` | Reads pages of an attached PDF, images and PDFs in the sandbox, and figures in hospital documents as images. Also used to check figures and PDFs the model has made |
| Hospital documents | `list_doc` / `grep_doc` / `read_doc` | Listing, search, and text retrieval of the document tree addressed by path. The source is the path |
| Skills | `load_skill` | Loads the body of a skill, of which only the description is always visible, when it is needed. The tools that come with the skill are added at that time |
| Code and files | `run_python` / `run_python_once` / `run_shell` | Python (state kept within a conversation) and shell in the sandbox. Plotting, statistics, and conversion |
| | `write_file` / `read_file` / `list_workspace` | Creating, viewing, and listing files in the workspace |
| | `send_to_workspace` | Takes attachments and original hospital documents into the workspace |
| | `fetch_from_workspace` | Hands a file in the workspace to the user. Images are displayed in place. Other files become a link the user can open |
| Subagents | `run_subagent` | Passes materials and instructions to a subagent and |

| Category | Tool | What it does |
|---|---|---|
| | | receives only its conclusion. To ask a follow-up question, the same session is specified. Available only when the skill `subagent` is loaded, and limited to administrators and designated groups |
| Background tasks | `check_task` / `await_task` / `stop_task` | Checks the status of, waits for, or stops time-consuming processes running in the background |
| Context budget | `get_context_budget` | Returns the remaining context. The model uses this to decide whether to read by itself, delegate to a subagent, or give its conclusion first |
| Conversation history | `grep_history` / `read_history` | Searches and reads the user's past conversations with the model. Facts the model is asked to keep are placed in `memory/` in the workspace |
| Speech | `transcribe_speech` / `inspect_audio` / `synthesize_speech` | Transcription of recordings, inspection of segments, and speech synthesis |

## Reading Materials

Attached Word, Excel, PowerPoint, PDF, and comma-separated values (CSV) files are converted to Markdown inside Nucllei by MarkItDown [28] (no optical character recognition [OCR] or layout analysis), and their text goes into the context as it is. Nucllei does not summarize or excerpt an attachment: the full text goes into the context, or, if it does not fit, nothing does. Instead, the model takes it into the sandbox to aggregate, extract, or split it, or has a subagent read it. We did not adopt the method of retrieving only part of an attachment by search and passing that part. This is to avoid the model answering as if it had read the whole document when it has read only part of it.

Images are not converted but given to the model directly as images. A PDF without a text layer (a scan, fax, or photograph) yields empty text extraction, so the model reads its pages as images. For forms with fields (consent forms, laboratory report sheets, referral documents, and application forms), the system prompt instructs the model to output what it reads in a fixed format that distinguishes printed text, handwriting, checkboxes, seal impressions, and blank fields, and not to fill unclear numbers or dates with plausible values.

## Hospital Documents

Hospital documents are placed as a tree of directories, one per domain. The leaves are units that answer one question in one read, such as a section of the fee schedule, a clinical question (CQ) of a guideline, a section of a manual, or the definition of a clinical scale, and each directory has an index node that states its scope and table of contents. At the time of the evaluation, the document collection had 6 domains (the medical fee schedule, the Diagnosis Procedure Combination [DPC], neurological disease guidelines, thoracic oncology guidelines, CTCAE v5.0 in the Japanese translation by the Japan Clinical Oncology Group [JCOG], and the hospital manuals and information system frequently asked questions [FAQ]) and consisted of 7,954 files including the indexes (1,450 for the hospital manuals and information system FAQ, 4,126 for the medical fee schedule and DPC, 2,377 for the clinical practice guidelines and CTCAE, and 1 top-level index). CTCAE was included under the JCOG terms of use, which permit its incorporation into medical information systems [29]. The document service reads this tree at startup and exposes it through 3 read-only tools: listing, search, and retrieval of the text. Search is string matching on the normalized text (regular expressions can also be given), and no vector index is built. When a search finds nothing, the model searches again with different words or starts over from an index node. The path of the unit that was read serves as its source, and answers include the path. The path is the same string as a tool argument, as a file in the sandbox, in the page the user opens, and in the static serving of originals and figures, so the model never has to convert a path into another form.

The same tree is also mounted read-only in the sandbox. Because the listing of the document tools does not recurse and the search returns only the top results, conditional extraction such as "all DPC classifications whose period II is 30 days or longer" and counting are done in the sandbox. The document tools are used for reading and quoting text.

## Skills

A skill is a guide provided by the developers. It holds procedures such as how to look up the documents of each domain, how to handle file conversion and large attachments, and how to calculate a clinical scale. Only the one-line description of each skill is visible to the model at all times, and the body is loaded when it is needed. The tools that come with a skill are added at that time. In Open WebUI, the user selects a prompt template and enables tools separately for each chat. In Nucllei, these two are combined into a single mechanism, the skill. At the time of submission, there are 21 skills, of which 13 are skills for the hospital documents (12 for single domains and 1 for counting and comparison across domains).

## Sandbox

The sandbox is an execution environment for Python and the shell, isolated with nsjail [30], and it cannot communicate with the outside or with the hospital network except through the APIs prepared in advance. Each user has a separate workspace (a directory) that persists across conversations. It holds the imported attachments, a folder per task, and the memory and recipes described below, and the group and global shared locations are mounted with permissions. The Python state is kept within a conversation. Because the sandbox cannot communicate with the outside, packages cannot be fetched on demand. The packages commonly used in administrative and analytical work were therefore bundled from the start: statistics (survival analysis, effect sizes, and baseline characteristics tables), plotting (with a Japanese font as the default), reading, writing, and generating Office documents and PDFs, legacy Word and Excel formats, Access, compressed files, OCR, Digital Imaging and Communications in Medicine (DICOM), Health Level Seven (HL7), and IBM SPSS Statistics files. The design does not require users to write code.

The model makes artifacts in the workspace and hands the finished ones to the user with a tool (`fetch_from_workspace`). Figures are displayed in place, and Excel and PDF files can be downloaded.

## Self-Checking of Generated Output

The model can read as images not only attached PDFs and the figures in hospital documents but also the figures and PDFs it has made itself in the sandbox. It checks a figure after plotting it and fixes overlapping or missing labels if there are any, and it can also check a PDF with a layout after making it and fix a broken layout. Because the same model has both image input and the sandbox, it can repeat making and checking without going through the user.

## Subagents

Understanding long materials and investigating several topics are delegated to subagents. The parent agent passes only the location of the materials and the instructions, and the subagent opens and reads them itself and returns only its conclusion. The content of the materials does not enter the context of the parent agent. The parent agent has a tool that measures its remaining context, and decides whether to read by itself, have a subagent read, or give its conclusion first. Because each subagent call adds a request to be processed on the shared GPUs, subagents are available only to administrators and designated groups, controlled through the visibility of the skill.

## Memory and Recipes

Memory and recipes are both plain mechanisms that sit on this workspace and only place Markdown files in it. Memory is a set of notes about the user (affiliation, preferences, work in progress), kept in the user's personal location as an index and one file per topic. It is written only on explicit request, read by the model when a request depends on the user's own context, and not shared. A recipe is a record of a procedure. After a task has gone well, when the user asks the model to "make this a recipe," the model writes out the steps, saves them, and shows the saved text. Whereas a skill is a guide to using the system, written by the developers and updated together with the harness, a recipe is a procedure for doing the work that a user saves as their own file and that users can share and improve. There are 3 locations, personal, group, and global, and where a recipe can be written is decided by access control: the sandbox's file permissions mirror the scope of users and groups in Nucllei (members can write to their group, and administrators to the global level).

## Speech

Microphone input is transcribed with Whisper large-v3-turbo [31] run by faster-whisper (CTranslate2, 8-bit integer [int8]) [32], and responses are read aloud with the AivisSpeech engine [33] and a publicly available Japanese voice model, both on the host's central processing units (CPUs). The browser's cloud speech recognition is not used, because the closed network cannot connect to the outside and patients' voices cannot be sent out. Attached recordings are transcribed before they are handled.

## Controls

Tools, skills, and presets are assigned on the server side, and what is visible is separated by group. Hospital documents are read-only, and neither users nor the model can modify them. The sandbox cannot communicate except through the APIs prepared in advance, and the workspaces of other users are not visible. The locations of recipes follow the same permissions. The possibility remains that the model follows instructions planted in an attachment (prompt injection). Even then, what can be changed without the user's intent is limited to that user's own conversation and workspace, and to the shared locations that user has permission to write to. Quotations of hospital documents remain as paths, and which tools were called and how is kept in the conversation record.

## Evaluation

### Access to Hospital Documents

For access to hospital documents, the accuracy of the configuration in which the model reads the documents itself (interactive access) was compared with fixed retrieval. The questions are 180 Japanese questions (6 domains × 30 questions) written by the development team, and the distribution of the 6 types (lookup, combined-source, filtering/counting, unavailable-source, figure/table, and edition identification) was decided before the questions were written (Multimedia Appendix 2). Each question has a reference answer that can be matched automatically, accepted variants, and the paths of the document units that support the answer (evidence paths), and the authors checked the reference answers against the original text. The first drafts of the questions and reference answers were made with Claude Code (Claude Fable 5.1 and Claude Opus 5.5; Anthropic) and reviewed and edited by the development team. The version of the document collection was fixed before the runs.

The 3 configurations compared were no documents, fixed retrieval, and interactive access. Fixed retrieval splits all files of the document collection into chunks of 2,000 characters (overlap 200), gives the top 10 chunks by Best Matching 25 (BM25; k1=1.2, b=0.75; after normalization to Unicode Normalization Form Compatibility Composition [NFKC], Japanese text is split into character bigrams and alphanumeric text into words) with the question text as the query once, and gives no orientation text or skills. Interactive access gives the document tools and the sandbox together with an orientation text for the document collection and 7 domain-specific skills (texts in Multimedia Appendix 3), and does not include subagents, image input, recipes, or memory. The sandbox-only configuration (which can read the document tree through the mount) and the document-tools-only configuration were measured in the same runs (Table S2). The model, the answer instructions, and the decoding settings (thinking on, temperature 1.0, top-p 0.95) were common, and each question was run 3 times with different request seeds (3 seeds below), with limits of 90 tool calls and 1,800 s per question.

Grading is automated matching of the final answer against the reference answer. Timeouts, reaching the call limit, and execution errors were counted as incorrect (execution failures below).

For fixed retrieval, whether the retrieved chunks included the evidence paths was also counted (Table S3). Differences in accuracy between configurations were evaluated on the per-question means over 3 seeds, with 95% confidence intervals (CIs) from a bootstrap (10,000 replicates) stratified by domain and with sign-flip tests. The main comparison is the difference between interactive access and fixed retrieval. Wall time, the number of calls, and the comparisons with sandbox only and document tools only were secondary and adjusted for multiplicity by the Holm method (Tables S2 and S4). Comparisons by question type are exploratory. As checks on the grading, the reference answers were checked against the document collection by script, and candidates for grading errors were extracted by pattern and read by the authors for a sensitivity analysis with corrected verdicts (Table S4).

### Automated Test of Component Triggering

The correspondence between scenarios and components ("Harness Components") was confirmed by an automated test on the harness as of submission (Multimedia Appendix 4). We wrote each of the 13 scenarios as one test, gave the request text and attachments to the model as a chat, and determined mechanically from the record of tool calls which of the 10 components were triggered (for example, the document tree by a call to a document tool, self-checking of generated output by a call to `feed_image` on an image the model made itself, and recipes by execution of a recipe command in the sandbox). The request texts are summaries of the requests used in the demonstrations for the 5 demonstrated scenarios, and 1 or 2 sentences that name the route that uses the component for the 8 envisioned scenarios (for example, "have subagents read it chapter by chapter" and "look up the defining formulas of the clinical scales in the hospital documents"). All request texts are in Multimedia Appendix 4. The attachments and preconditions (laboratory values, scans without a text layer, a meeting recording, files in legacy formats, recipes, and memory) were all synthesized. Each scenario was run 4 times, and we counted the number of runs in which the expected components were triggered. The correctness of the answers was not judged.

### Scenarios on the Deployment Host

The 5 demonstrated scenarios were run once each by the authors on the deployment host with synthetic data, and we recorded the input, the order of the components used, and the time taken. The user interface (UI) and skills were those of the deployment, and serving used SGLang 0.5.15.post1, an earlier version than in the serving tests. The transcription of the referral letter was checked against the synthetic original.

## Ethical Considerations

This study does not involve human participants. It uses no data, samples, or information about people, including patients and staff (including the usage records since the platform went into operation), and involves no intervention or observation. The data used in the evaluation are limited to public benchmarks, hospital documents and published guidelines, synthetic data, and questions written by the development team. The study therefore falls outside the scope of the Ethical Guidelines for Medical and Biological Research Involving Human Subjects (Ministry of Education, Culture, Sports, Science and Technology; Ministry of Health, Labour and Welfare; and Ministry of Economy, Trade and Industry) [34] and is not subject to ethics review. Operation of the platform is part of the hospital's information system operations and is outside the scope of this study.

# Results

## Operation

The platform was built and tested on the university network, moved to the hospital's medical information network on September 14, 2026, and has been in operation since. The measurements and demonstrations in this paper were made on the same host before the move. The platform has continued to be updated since the move, and the harness described in this paper is the one as of submission (October 2026).

## Model and Serving Performance

Table 3 shows the benchmark scores of the 3 candidate models with and without thinking (the breakdown for text-only questions and the IgakuQA scores by year are in Table S1 in Multimedia Appendix 1). On the 2026 Japanese Medical Licensing Examination (with thinking), Kimi K2.6 scored 480/500 (with images), with 191/200 on the compulsory section and 289/300 on the general and clinical sections, above the score criteria for passing (160/200 and 224/300). On the public leaderboard that uses the same scoring, its score is 13 points below the top cloud models (Claude Opus 4.6 and Gemini 3.1 Pro, 493/500) [18]. The differences among the 3 candidates were small: on the text-only questions, where the inputs were matched, the 2026 scores were 369 for Kimi K2.6 and 370 for GLM-5.1 and DeepSeek V3.2, a difference of 1 point (Table S1). Without thinking, Kimi K2.6 ranked first in both years, on all questions and on text-only questions alike. On the secondary benchmarks JMED-LLM (3 tasks) and IgakuQA 2018-2022, with thinking, Kimi K2.6 and GLM-5.1 were the top 2 and DeepSeek V3.2 was third. The scores on the 4 tasks of llm-jp-eval are in Table 3 (B). Kimi K2.6 was selected because its scores were comparable to or better than those of the other 2 candidates and because it keeps image input and a long context while leaving enough KV cache for concurrent requests (Table 1). Image input is a prerequisite for the component that reads scanned forms and for the component by which the model checks its own output. In public evaluations by third parties under common protocols, Kimi K2.6 was on par with GLM-5.1 and above DeepSeek V3.2 on tool use, the lowest of the 3 on a search-agent evaluation, and the highest on long-context reasoning (Table S5).

The summed decode speed of the requests processed concurrently (aggregate output throughput) was 908 tokens/s at 16 concurrent requests with EAGLE-3 (Table 4). For a single user, generation ran at 119 tokens/s with thinking and 129 tokens/s without thinking, with EAGLE-3. In the input-length sweep, the time to first token was 0.26 s at 1,130 input tokens and 41.6 s at 245,891 input tokens. Mean GPU power (total of the 8 GPUs) was about 760 W with the model loaded and idle and 2,337 W under a load of 16 concurrent requests (a separate run from Table 4).

Table 3. Benchmark scores of the candidate models (our local runs). In the examinations, only Kimi K2.6 received the question images (text-only scores and IgakuQA by year in Table S1). In MGSM-ja with thinking, GLM-5.1 gave no \boxed{} required by the grader in 127 of the 250 questions, which were counted as wrong (in 115 of them, the last number matched the reference).

**(A) Medical benchmarks**

| Model | Thinking | 2026 examination (JMLE2026), /500 | 2025 examination (IgakuQA119), /500 | IgakuQA 2018-2022, /2,485 (%) | JMMLU-Med, κ (accuracy) | CRADE, weighted κ (accuracy) | RRTNM, κ (accuracy) |
|---|---|---|---|---|---|---|---|
| Kimi K2.6 | on | 480 | 474 | 2,245 (90.34) | 0.90 (0.92) | 0.67 (0.81) | 0.90 (0.93) |
| GLM-5.1 | on | 481 | 478 | 2,283 (91.87) | 0.89 (0.92) | 0.64 (0.81) | 0.89 (0.92) |
| DeepSeek V3.2 | on | 475 | 478 | 2,205 (88.73) | 0.86 (0.89) | 0.55 (0.70) | 0.86 (0.89) |
| Kimi K2.6 | off | 472 | 453 | 2,113 (85.03) | 0.85 (0.89) | 0.57 (0.73) | 0.84 (0.88) |
| GLM-5.1 | off | 456 | 442 | 2,129 (85.67) | 0.86 (0.89) | 0.50 (0.65) | 0.77 (0.83) |
| DeepSeek V3.2 | off | 458 | 440 | 2,053 (82.62) | 0.83 (0.87) | 0.58 (0.78) | 0.74 (0.81) |

**(B) llm-jp-eval**

| Model | Thinking | JCommonsenseQA, EM | JEMHopQA, EM / character F1 | JSQuAD, EM / character F1 | MGSM-ja, accuracy |
|---|---|---|---|---|---|
| Kimi K2.6 | on | 0.979 | 0.617 / 0.747 | 0.806 / 0.912 | 0.904 |
| GLM-5.1 | on | 0.977 | 0.658 / 0.764 | 0.812 / 0.921 | 0.432 |
| DeepSeek V3.2 | on | 0.979 | 0.550 / 0.664 | 0.801 / 0.912 | 0.884 |
| Kimi K2.6 | off | 0.963 | 0.325 / 0.357 | 0.817 / 0.925 | 0.880 |
| GLM-5.1 | off | 0.949 | 0.367 / 0.395 | 0.634 / 0.724 | 0.808 |
| DeepSeek V3.2 | off | 0.920 | 0.300 / 0.351 | 0.815 / 0.920 | 0.872 |

EM: exact match; F1: character-level F1 score.

Table 4. Serving performance of Kimi K2.6 in the deployment configuration (conditions in Methods, “Model and Serving”).

**(A) Aggregate output throughput under concurrent requests**

| Concurrent requests | EAGLE-3 on, tokens/s | EAGLE-3 off, tokens/s | Ratio | TTFT (EAGLE-3 on), median / p99, ms | Mean accepted draft length |
|---|---|---|---|---|---|
| 4 | 544 | 231 | 2.4 | 73 / 119 | 2.76 |
| 8 | 726 | 396 | 1.8 | 125 / 276 | 2.91 |
| 16 | 908 | 652 | 1.4 | 157 / 349 | 3.04 |

**(B) Single-user decode speed**

| Mode | TTFT, ms | Decode speed (EAGLE-3 on, mean of 3), tokens/s | Decode speed (EAGLE-3 off), tokens/s |
|---|---|---|---|
| Thinking off | 72-93 | 129 | 80 |
| Thinking on | 71-72 | 119 | 80 |

TTFT: time to first token; p99: 99th percentile; mean accepted draft length: mean number of tokens accepted per EAGLE-3 verification step (the accept length reported by SGLang).

## Interactive Access Compared With Fixed Retrieval

Mean accuracy was 10.7% with no documents, 81.3% with fixed retrieval, and 97.2% with interactive access. The number of questions answered correctly in all 3 seeds was 143 with fixed retrieval and 169 with interactive access. Interactive access exceeded fixed retrieval by 15.9 percentage points (95% CI 10.9-21.3; *P*<.001). The conclusion did not change after the 18 wrong verdicts of the automated grader were corrected (16.9 percentage points, 95% CI 11.5-22.4; Table S4).

The difference was concentrated in the filtering/counting type (Figure 2). On this type, fixed retrieval scored 16.7% and interactive access 87.8%, a difference of 71.1 percentage points (95% CI 57.8-83.3; *P*<.001). On the lookup type, in contrast, fixed retrieval scored 96.3% and interactive access 100% (difference 3.7 percentage points; *P*=.13), on the combined-source type 91.5% and 98.3% (*P*=.25), and on the edition identification type fixed retrieval also answered every question correctly. Among the questions that have evidence paths, the proportion of responses in which the chunks retrieved by fixed retrieval included at least one of the question's evidence paths was 93.5% overall and 65.4% for the filtering/counting type, and no response whose chunks included none of the evidence paths was correct. On the filtering/counting type, none of the 21 responses whose chunks included only some of the evidence paths was correct, and of the 32 responses whose chunks included every evidence path, 12 were correct (Table S3). Almost all correct answers with no documents were on the unavailable-source type, where the model answered that the information was unavailable. Interactive access had no execution failures, and the mean wall time per question was 26.2 s, with a mean of 7.0 tool calls.

On the same question set, sandbox only and document tools only scored 97.0% and 94.6%, respectively, and the differences from interactive access (both) were 0.2 percentage points (Holm-adjusted *P*>.99) and 2.6 percentage points (95% CI 0.7-4.6; Holm-adjusted *P*=.04, and $P$=.07 after grading correction) (Tables S2 and S4). The difference lay in the filtering/counting type, where sandbox only scored 91.1%, document tools only 74.4%, and both 87.8%, and with document tools only, 6 responses failed (3 reached the 1,800-s limit and 3 ended with a connection error). On the lookup type, where accuracy was close to 100%, interactive access (both) answered in 13.3 s and 4.5 calls per question, whereas sandbox only took 21.6 s and 6.1 calls (difference −8.3 s, 95% CI −9.8 to −6.9; Holm-adjusted $P$<.001; Table S2).

In the scripted check of the reference answers, 161 of the 180 questions were verified by a quotation from the original text, a recalculation from a table, or a search confirming absence. For the other 19, the quoted evidence was a paraphrase, and we confirmed the answer at the referenced location. No reference answer contradicted the document collection. One reference answer was correct with respect to the document collection but reflected an ingestion error made during conversion, which was corrected in the collection after the evaluation.

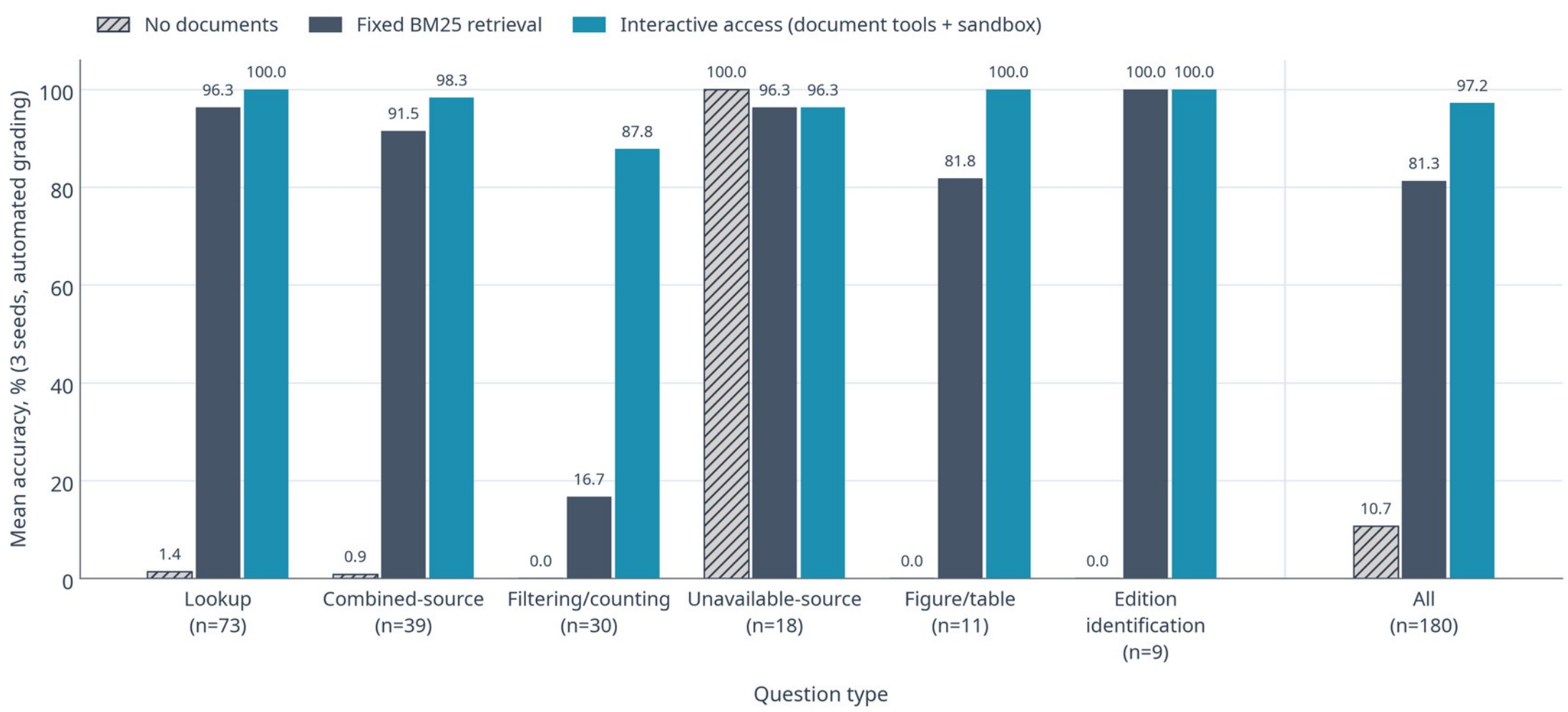

*Figure 2. Mean accuracy by question type (3 seeds, automated grading) for no documents, fixed BM25 retrieval, and interactive access (both document tools and sandbox). The values for the sandbox-only and document-tools-only configurations are in Table S2. Values by question type are exploratory, and intervals were computed only for differences between configurations (Table S4). No documents reaches 100% on the unavailable-source type because that configuration answers that the information is unavailable.*

## Scenarios and Components

Table 5 shows the 13 scenarios and the components each uses. All scenarios except scanned referral letter transcription need more than one component. In batch CTCAE grading, the model reads the attached laboratory values in full, looks up the text of the criteria in the document tree, and builds a grade table that quotes the criteria. The listing and search of the document tools alone can reach the criteria, but a skill gives prior knowledge of how to search by organ system and of how the documents are organized, which is intended to shorten the search. Aggregating falls data can be carried out only with the functions that pass the attachment to the sandbox, process it with the statistics and plotting packages, and return the figure to the user.

In the automated test (Methods, “Automated Test of Component Triggering”), the components expected in Table 5 were triggered in 4 of 4 runs for 12 of the 13 scenarios. In the remaining scenario, falls data aggregation and plotting, the full text of the attachment and the sandbox were triggered in all 4 runs, but self-checking in 3 and returning in 2, with runs in which the model handed over the figure without looking at it and runs in which it looked at the figure but did not hand it over (Multimedia Appendix 4).

Table 5. Work scenarios and the harness components needed to carry them out. The upper 5 rows are the scenarios demonstrated on the deployment host (Results, “Scenarios on the Deployment Host”), and the lower rows are envisioned scenarios that can be built from the same components (Methods, “Harness Components”). The columns correspond to components (1) to (10) in Methods, “Harness Components”. Speech input and output are not included as a column. Each ✓ was triggered in 4 of 4 runs in the automated test of the harness as of submission (Methods, “Automated Test of Component Triggering”). The exceptions are Self-check (3/4) and Return (2/4) in falls data aggregation and plotting (Multimedia Appendix 4).

| Scenario | Page images | Attachment full text | Document tree | Sandbox | Return | Self-check | Subagents | Skills | Recipes | Memory |
|---|---|---|---|---|---|---|---|---|---|---|
| **Demonstrated** | | | | | | | | | | |
| Needlestick injury response | | | ✓ | | | | | ✓ | | |
| Batch CTCAE grading | | ✓ | ✓ | | | | | ✓ | | |
| Scanned referral letter transcription | ✓ | | | | | | | | | |
| Falls data aggregation and plotting | | ✓ | | ✓ | ✓ | ✓ | | | | |
| Investigation of 3 topics by subagents | | | ✓ | | | | ✓ | ✓ | | |
| **Envisioned** | | | | | | | | | | |
| Research data analysis | | ✓ | | ✓ | ✓ | ✓ | | ✓ | ✓ | ✓ |
| Lab report transcription and trend chart | ✓ | | | ✓ | ✓ | ✓ | | | | |
| Legacy format conversion | | ✓ | | ✓ | ✓ | | | ✓ | | |
| Clinical scale calculation | | | ✓ | ✓ | | | | ✓ | | |
| Billing requirement check | | | ✓ | ✓ | | | | ✓ | | |
| Summarizing long materials | | ✓ | ✓ | ✓ | | | ✓ | ✓ | | |

| Scenario | Page images | Attachment full text | Document tree | Sandbox | Return | Self-check | Subagents | Skills | Recipes | Memory |
|---|---|---|---|---|---|---|---|---|---|---|
| Minutes from a recording | | | | ✓ | ✓ | | | ✓ | | |
| Sharing and rerunning procedures | | | | ✓ | ✓ | | | | ✓ | ✓ |

## Scenarios on the Deployment Host

Table 6 shows the input, the order of the components used, and the time taken for the 5 scenarios demonstrated on the deployment host. The time taken ranged from 19.7 s for the needlestick injury inquiry to 289.6 s for the investigation of 3 topics. For the scanned referral letter without a text layer, the model read the page image, structured the patient information, diagnoses, history, and 7 laboratory items, and pointed out 5 items outside the reference ranges. The transcription had no errors when checked against the synthetic original. In the investigation of 3 topics, the parent agent started 3 subagents one after another (41.3 s, 187.9 s, and 31.4 s) and merged the conclusions each returned from reading the hospital manuals into a comparison table. Figure 3 shows the screen for one of these, the batch CTCAE grading.

Table 6. The 5 scenarios demonstrated on the deployment host (Methods, "Scenarios on the Deployment Host"). The time taken is the sum of the model generation time recorded by Nucllei (cumulative over the turn) and the tool execution time. The requests were entered in Japanese and are given here as summaries in English translation.

| Scenario | Input | Components used (in order) | Time taken (generation + tool execution) |
|---|---|---|---|
| Needlestick injury response | "I pricked my finger with a used needle after drawing blood. Following our hospital's manual, tell me what to do, starting with what to do right now." | Infection control skill → `read_doc` on the relevant section of the hospital manual → procedure and contacts quoted verbatim | 19.7 s (generation 19.7, tools 0.0) |
| Batch CTCAE grading | Attached a list of laboratory values and symptoms for 6 patients after chemotherapy (CSV): "Grade them by CTCAE v5.0 and quote the text of the | CTCAE skill → `grep_doc` / `read_doc` for each organ system → grade table of patients × items and quotation of each criterion | 86.7 s (generation 86.6, tools 0.1) |

| Scenario | Input | Components used (in order) | Time taken (generation + tool execution) |
|---|---|---|---|
| | criteria." | | |
| Scanned referral letter transcription | Attached a scanned image of a referral letter without a text layer: "Structure it for transcription and point out any values outside the reference ranges." | Read the page image → structured the patient information, diagnoses, history, and 7 laboratory items → pointed out 5 items outside the reference ranges | 36.5 s (generation 36.5, no tools) |
| Falls data aggregation and plotting | Attached 12 months of falls data by ward (CSV): "Calculate the incidence rate and plot the monthly trend, check any increasing trend statistically, and give the key points in 3 lines." | Aggregation and regression in the sandbox → generated a figure with Japanese labels → figure displayed inline, key points in 3 lines | 38.8 s (generation 37.8, sandbox 1.1) |
| Investigation of 3 topics by subagents | "Look up the 3 topics of needlestick response, falls response, and incident report submission in parallel and put them in a comparison table." | Started 3 subagents one after another (41.3 / 187.9 / 31.4 s) → each loaded a skill, read the hospital manuals, and returned its conclusion → parent agent merged them into a comparison table | 289.6 s (generation 29.0, subagents 41.3 + 187.9 + 31.4) |

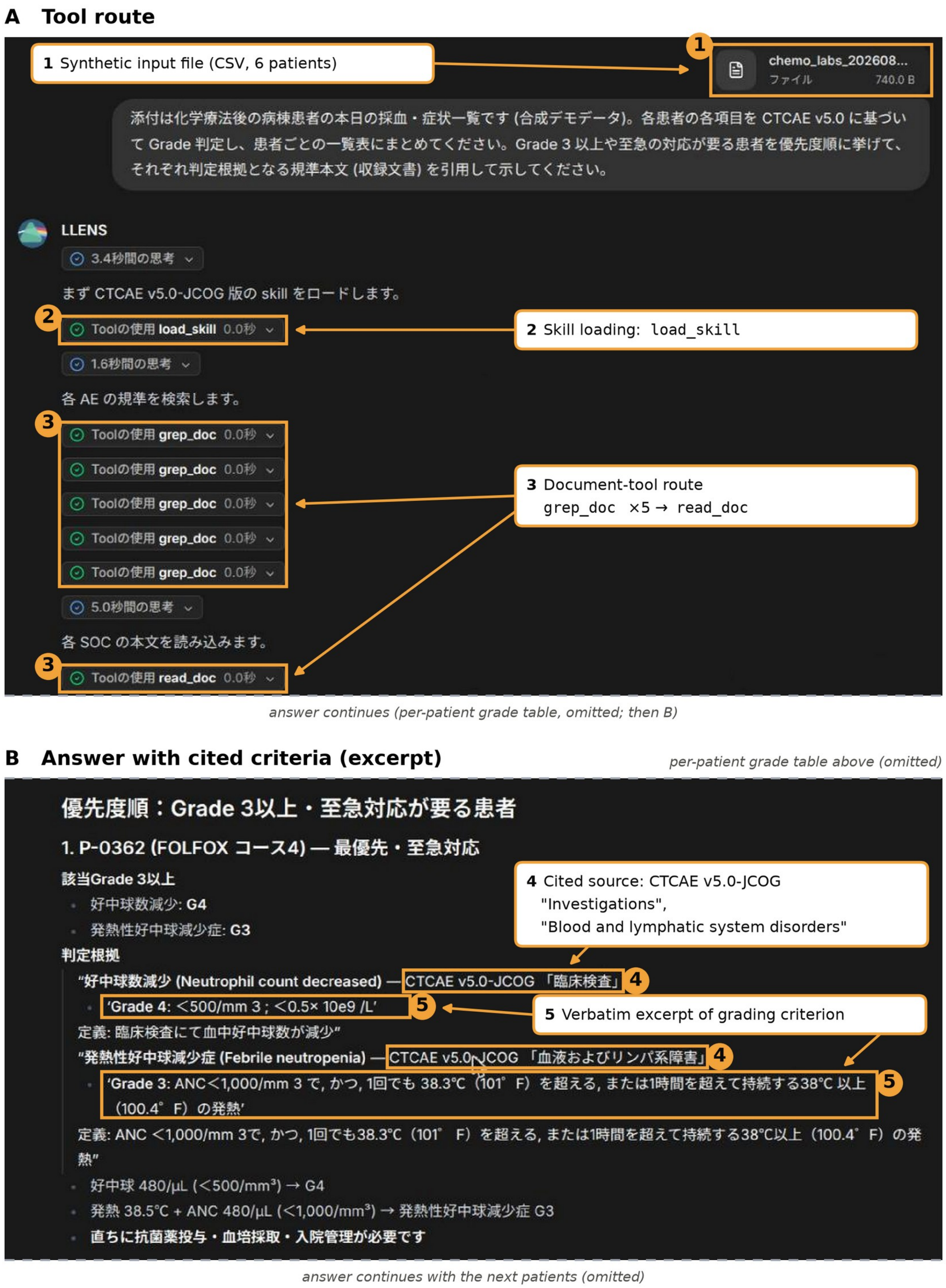

*Figure 3. Example LLENS session in which synthetic laboratory findings were graded by CTCAE v5.0 with the criteria quoted. The user attached a synthetic CSV file of laboratory values and symptoms for 6 ward patients after chemotherapy and asked for grading. The model loaded the skill, searched and read the criteria with the document tools (A), and answered with the source units and verbatim criteria for the first patient in the requested priority order (B). The UI*

*is in Japanese. The seconds next to the tool names are tool execution times (under 0.1 s for the document tools), and most of the time taken is the model's generation.*

# Discussion

## Principal Findings

An agent platform with an open-weight model and the harness on a single host carried out multistep work, such as consulting hospital documents, reading scanned forms, aggregating and plotting attached data, and dividing an investigation among subagents. The platform is in operation inside the closed network. Running the components requires conditions of the model beyond examination scores. Reading forms and the figures the model itself made requires image input, receiving attachments in full requires a long context, choosing and calling tools for each request requires native function calling, and use by several staff members at once requires room in the KV cache. Kimi K2.6 met these conditions within one host, and the GPU power of that host was 2.3 kW even under a load of 16 concurrent requests.

Materials are read by the model itself rather than passed as fragments extracted by retrieval (an attachment goes in either in full or not at all, and the hospital documents form a tree addressed by path). Reports that keyword search over files by an agent approaches vector search [35,36] support this choice. As a result, whether the model has read a material, and its source, are clear. A revision is a file replacement, and no separate database has to be brought into the closed network. In the evaluation, fixed retrieval, which passes the top 10 chunks from a single search with the question text as the query, scored only 16.7% on the filtering/counting type. Of its 81 responses to the 27 questions of this type that have evidence paths, the 49 whose chunks lacked some or all of the evidence paths were all incorrect, and only 12 of the 32 (37.5%) whose chunks included every evidence path were correct (Table S3). A single search often fails to retrieve the documents an answer needs, and even when it does, the top 10 chunks alone are often not enough to build the answer. When the model decided for itself what to read with the document tools and the sandbox, it reached 87.8% on the same type. In the configurations limited to one kind of tool, the sandbox alone gave almost the same accuracy, but lookup questions were answered faster with both, and the document tools alone fell on the filtering/counting type. This matches the division of roles intended in the harness: the faster document tools for reading and citing text and the sandbox for listing and counting the matching documents.

## Comparison With Prior Work

Earlier reports of running a model inside a hospital are each built for one use: extracting findings from medical histories with a small open-weight model [5], adding retrieval-augmented generation to contrast media consultations [8], combining verified fixed texts with generated small talk in preoperative patient interaction [37], and extracting items from pathology reports [9]. Agents deployed in hospitals are likewise fixed to drafting specific documents: emergency department discharge summaries and preanesthesia evaluations within the electronic health record [10] or hospital course summaries [11]. Agents at the research stage remain evaluations on fixed tasks: questions over structured records [3], reasoning through discussion among agents with separate roles [38], and oncology decisions with tools [4]. In LLENS, each component is an existing technique, but both the demonstrated and the envisioned scenarios are assembled from combinations of the same components, and new work is covered by adding components and procedures. Procedures are divided into how to use the system, written by the developers (skills), and how to do routine work, saved by users (recipes). Because users save recipes and share them within a department or across the hospital, they can widen the uses of the platform without going through the developers.

What the model reads and returns has also been decided by the system. Retrieval-augmented generation in medicine has been evaluated as passing chunks chosen by a retriever, whether on public examination questions [39], a curated guideline resource [40], or a hospital's antimicrobial guidelines [41]. The platform of Grünig et al, a general-purpose in-hospital LLM for all staff, was limited to prompting and an API, and its users asked for PDF input and export to Excel [12]. In LLENS, staff pass materials from the browser, the model chooses what to read and which tools to use for each request, and artifacts are returned as files.

## Next Steps

Going forward, we will add to the harness a function that connects directly to the electronic health record database, through Structured Query Language (SQL) or other means. At present, users have to export data from the electronic health record and attach it. Once the connection is in place, in writing a summary for an inpatient, for example, a physician will specify the patient and make the request, and the model will retrieve the progress notes, laboratory results, and prescriptions from the electronic health record and prepare a draft in the hospital's format. The physician then checks and corrects it. Reading materials, aggregating data, and preparing documents in a required format can already be done with the current components, so what is newly needed is a function that retrieves data from the electronic health record. After the electronic health record, we plan to connect departmental systems such as those for laboratory, imaging, pathology, and pharmacy in the same way. The aim is to reduce the time physicians spend on the electronic health record and desk work and let them use it for patient care and research.

The evaluation of work and users will be done in the workplace in parallel with the connection to the electronic health record, and it will observe how the workplace changes after AI agents are introduced. It will start with the time taken by routine work, which is easy to measure, by comparing the input time for case registration before and after the connection is introduced. It will then observe user behavior and behavior change, and widen further to the quality of care and patient outcomes.

## Limitations

This paper is limited to the description of the platform, the technical evaluation of the model and of access to hospital documents, and demonstrations with synthetic data, and it does not include evaluation in actual clinical settings (usage records, how users receive the platform, effects on work, and patient outcomes). Measuring work handles data in the electronic health record, and measuring user behavior is research on hospital staff, so these will be done separately after ethics review (Next Steps). Each scenario was demonstrated in a single run, and the automated test confirmed whether the expected components are triggered by the request texts of the test. Neither is an evaluation of the accuracy or reproducibility of the answers. The scenarios and document access were evaluated only with the selected model and were not compared with models of different size or capability, so how far these results depend on the model has not been shown. The questions for the comparison of document access were written by the team that built the document tree and the skills, with the distribution of types decided in advance, so the overall difference depends on the mix of types. The comparison was made against untuned fixed retrieval and is not a comparison at an equal inference budget with retrievers that include vector search or query rewriting. The training-data cutoffs of the candidate models are not disclosed, and their exposure to the licensing examination cannot be verified.

## Conclusions

We built LLENS, a clinical support AI agent platform that runs on one host inside a hospital's closed network with a trillion-parameter-class open-weight model and an agent harness, and it is in operation. The model carried out multistep work that combines reading materials itself, computing, and returning artifacts, and with the hospital documents it chose what to read and answered questions that require filtering or counting across several documents. Its operation shows that a hospital that cannot send patient information outside can host a general-purpose agent platform without building a system for each use. Next, we will connect the electronic health record and departmental systems so that the model reads patient records directly.

# Acknowledgments

We thank all members of PRISM-HU who took part in the development and research [13], and the staff of the Faculty of Medicine, Hokkaido University, and Hokkaido University Hospital for their support in building and deploying the platform. We thank the Japan Agency for Medical Research and Development (AMED) for selecting PRISM-HU for its Program for Supporting Medical Research [14]. We thank Naoto Iwase, a medical student who develops and maintains the JMLE2026 and IgakuQA119 benchmarks used in the model evaluation, and the developer communities of SGLang, Open WebUI, and Kimi K2.6.

# Funding

This work was supported by AMED (PRISM-HU) under Grant Numbers JP256f0137006 and JP266f0137006 [13], and by the Japan Society for the Promotion of Science (JSPS) KAKENHI Grant Number JP26K25020.

# Conflicts of Interest

KE is the founder and representative director of Technoplasm Inc., which co-developed the web application Nucllei, and holds equity in the company. The other authors declare no conflicts of interest.

# Data Availability

The source of the platform is published in 2 GitHub repositories. The web application Nucllei (technoplasm/nucllei) is published under terms based on the PolyForm Noncommercial License 1.0.0 [27]. The LLENS platform and the evaluation code as of the evaluation are published as a reference snapshot (prism-hu/llens-snapshot) [42]. The snapshot is for reading. It does not run without the services inside the hospital and the hospital documents.

The 180 evaluation questions (domain, question type, reference answer, accepted variants, evidence paths) are given in Multimedia Appendix 2, and the answer instructions, the orientation text for the document collection, and the text of the evaluation skills given to the model are given in Multimedia Appendix 3. The same content is also placed in the snapshot in other formats (jsonl for the questions, and Markdown for the answer instructions, orientation text, and skills). The public benchmark datasets are available from their providers under their terms. The document collection is not published (the hospital manuals and information system FAQ are internal documents, and the originals of the fee schedule, DPC, the guidelines, and CTCAE are publicly available from their issuers), and an end-to-end reproduction including tool execution requires the document collection managed by the hospital. This collection cannot be provided outside the hospital, and inquiries can be sent to the corresponding author.

# Author Contributions

KE: conceptualization, methodology, software, investigation, formal analysis, visualization, and writing (original draft). JS: validation of the platform in practical scenarios and proposals on the harness design and user experience. MM: methodology (input on design and operation). FH: software (supporting). ET: conceptualization (supporting). TK, RU, KS, MN, TT, and AE: resources and project administration (deployment, network, and operations as the Division of Medical Information Planning). KH and AY: funding acquisition and project administration. KK and ST: funding acquisition and supervision. All authors reviewed and approved the manuscript.

# Abbreviations

AI: artificial intelligence; AMED: Japan Agency for Medical Research and Development; API: application programming interface; BM25: Best Matching 25 (Okapi BM25 lexical ranking function); CI: confidence interval; CPU: central processing unit; CQ: clinical question; CSV: comma-separated values; CTCAE: Common Terminology Criteria for Adverse Events; DICOM: Digital Imaging and Communications in Medicine; DPC: Diagnosis Procedure Combination; DX: digital transformation; EHR: electronic health record; EM: exact match; F1: character-level F1 score; FAQ: frequently asked questions; FP8: 8-bit floating point; GPU: graphics processing unit; HL7: Health Level Seven; INT4: 4-bit integer; int8: 8-bit integer; JCOG: Japan Clinical Oncology Group; JMLE: Japanese Medical Licensing Examination; JSPS: Japan Society for the Promotion of Science; KV: key-value; LLENS: Large Language Enhanced Nexus System; LLM: large language model; MoE: mixture of experts; MTP: multi-token prediction; NFKC: Normalization Form Compatibility Composition (Unicode Normalization Form KC); OCR: optical character recognition; p99: 99th percentile; PDF: Portable Document Format; PRISM-HU: Strategic Promotion of Next-Generation AI-Enabled, Data-Driven, and Information-Circulating Medical Research through the Integrated Development of People, AI/DX, and Systems (Hokkaido University); SQL: Structured Query Language; TTFT: time to first token; UI: user interface.

# Multimedia Appendix

Multimedia Appendix 1: Supplementary tables (Table S1, breakdown of the candidate models' benchmark scores [scores and numbers correct on all and text-only questions of the Japanese Medical Licensing Examinations, and IgakuQA 2018-2022 by year]; Table S2, accuracy, wall time, and tool calls by question type for all 5 configurations; Table S3, coverage of the evidence paths by the chunks retrieved by fixed retrieval, and correctness; Table S4, sensitivity of accuracy and between-configuration differences to grading correction; Table S5, third-party public evaluations of the candidate models on tool use and long-context reasoning).

Multimedia Appendix 2: The 180 evaluation questions (xlsx; the questions, reference answers, accepted variants, and evidence paths are in Japanese as given to the model, with English column headers and labels).

Multimedia Appendix 3: Prompts and evaluation skills used in the evaluation.

Multimedia Appendix 4: Automated test of the triggering of the components in Table 5 (the request texts and preconditions of the 13 scenarios, the detection rules for the components, and the number of runs in which each component was triggered, by scenario × component).

# Multimedia Appendix 1. Supplementary tables

## Abbreviations

API: application programming interface; avg@4: accuracy averaged over 4 independent runs; BM25: Best Matching 25 (Okapi BM25 lexical ranking function); CI: confidence interval; CQ: clinical question; INT4: 4-bit integer; SE: standard error.

Table S1. Breakdown of the candidate models' benchmark scores with and without thinking (run locally, 1 response per question, temperature 0; the totals are the values in Table 3 [A]). (A) The Japanese Medical Licensing Examinations, the 120th (2026, JMLE2026) and the 119th (2025, IgakuQA119) [1,2]. Each cell is the score (number correct). The score weights each question by the benchmark's rules: all questions are 400 questions for 500 points, and the text-only questions are 302 questions for 382 points in 2026 and 297 questions for 383 points in 2025. Kimi K2.6 received the question images, whereas GLM-5.1 and DeepSeek V3.2 received only the text even for questions with images, so the inputs in the all-questions columns are not matched. The comparison with matched inputs is in the text-only columns. The 2025 values are rescored (Methods, "Model and Serving"). (B) IgakuQA 2018-2022 [3] by year: 5 examinations, 2,000 questions, 2,485 points, with questions that have images answered from text only. The total column gives the proportion of the maximum in parentheses (%).

### (A) Japanese Medical Licensing Examinations

| Model | Thinking | 2026, all questions | 2026, text-only questions | 2025, all questions | 2025, text-only questions |
|---|---|---|---|---|---|
| Kimi K2.6 (with images) | on | 480 (384) | 369 (291) | 474 (378) | 366 (284) |
| GLM-5.1 | on | 481 (383) | 370 (292) | 478 (382) | 370 (288) |
| DeepSeek V3.2 | on | 475 (383) | 370 (294) | 478 (380) | 372 (288) |
| Kimi K2.6 (with images) | off | 472 (374) | 362 (284) | 453 (361) | 348 (270) |
| GLM-5.1 | off | 456 (362) | 349 (273) | 442 (356) | 345 (271) |
| DeepSeek V3.2 | off | 458 (364) | 358 (280) | 440 (356) | 341 (269) |

### (B) IgakuQA 2018-2022 (text only)

| Model | Thinking | Total, points /2,485 (%) | 2018 | 2019 | 2020 | 2021 | 2022 |
|---|---|---|---|---|---|---|---|
| Kimi K2.6 | on | 2,245 (90.34) | 441 | 454 | 450 | 449 | 451 |
| GLM-5.1 | on | 2,283 (91.87) | 455 | 458 | 460 | 450 | 460 |
| DeepSeek V3.2 | on | 2,205 (88.73) | 437 | 441 | 448 | 434 | 445 |
| Kimi K2.6 | off | 2,113 (85.03) | 424 | 415 | 416 | 432 | 426 |
| GLM-5.1 | off | 2,129 (85.67) | 424 | 428 | 422 | 416 | 439 |
| DeepSeek V3.2 | off | 2,053 (82.62) | 406 | 412 | 401 | 414 | 420 |

Table S2. Accuracy, wall time, and tool calls by question type for all 5 evaluated configurations (the same 180 questions × 3 seeds), including sandbox only and document tools only. (A) Accuracy is the mean over 3 seeds (automated grading; grading correction does not change the comparison with fixed retrieval for any question type [Table S4]). No documents answers that the information is unavailable, so it is correct almost only for the unavailable-source type (the few other correct verdicts were grading errors; Table S4). Evidence-path recall of fixed retrieval is, over the questions that have evidence paths, the proportion of responses in which the chunks retrieved by fixed retrieval contained at least one of the question's evidence paths (the chunk's path equals the evidence path or lies under it). Unavailable-source questions have no evidence paths. No response that contained none of the evidence paths was correct. Execution failures are the number of failures among the 3 × n responses of sandbox only, document tools only, and interactive access (all counted as incorrect). (B) Wall time per question (s) is the 3-seed mean of each question averaged over questions, excluding questions with unmeasured time in any of sandbox only, document tools only, or interactive access (n measured). (C) Recorded tool calls per question is the per-question mean averaged over questions (median of the per-question means in parentheses). Each contrast is the difference (95% CI); unadjusted P; Holm-adjusted P. Intervals are pointwise 95% intervals of a question-level bootstrap stratified by domain († marks unstratified intervals). P values are from a paired sign-flip test. Holm adjustment is applied within the family of interactive access − sandbox only / interactive access − document tools only / document tools only − sandbox only for each question type and metric, not across question types or metrics. All analyses by question type are exploratory.

## (A) Accuracy, %

| Question type | n | No documents | Fixed retrieval | Sandbox only | Document tools only | Interactive access (both) | Evidence-path recall of fixed retrieval, % | Execution failures (sandbox only / document tools only / interactive access) |
|---|---|---|---|---|---|---|---|---|
| Lookup | 73 | 1.4 | 96.3 | 98.6 | 99.5 | 100.0 | 98.6 | 0 / 0 / 0 |
| Combined-source | 39 | 0.9 | 91.5 | 98.3 | 96.6 | 98.3 | 100.0 | 0 / 0 / 0 |
| Filtering/counting | 30 | 0.0 | 16.7 | 91.1 | 74.4 | 87.8 | 65.4 | 0 / 6 / 0 |
| Unavailable-source | 18 | 100.0 | 96.3 | 94.4 | 98.1 | 96.3 | — | 0 / 0 / 0 |
| Figure/table | 11 | 0.0 | 81.8 | 100.0 | 100.0 | 100.0 | 100.0 | 0 / 0 / 0 |
| Edition identification | 9 | 0.0 | 100.0 | 100.0 | 100.0 | 100.0 | 100.0 | 0 / 0 / 0 |
| All | 180 | 10.7 | 81.3 | 97.0 | 94.6 | 97.2 | 93.5 | 0 / 6 / 0 |

## (B) Wall time per question, s (mean)

| Question type | n (measured) | Sandbox only | Document tools only | Interactive access | Interactive access − sandbox only (95% CI) | Holm *P* | Interactive access − document tools only (95% CI) | Holm *P* |
|---|---|---|---|---|---|---|---|---|
| Lookup | 73 | 21.6 | 12.9 | 13.3 | −8.3 (−9.8 to −6.9) | <.001 | 0.4 (−0.3 to 1.2) | .38 |
| Combined-source | 39 | 30.4 | 19.0 | 19.0 | −11.3 (−15.0 to −8.0) | <.001 | 0.0 (−1.7 to 1.6) | .96 |
| Filtering/counting | 27 | 74.4 | 87.0 | 64.2 | −10.2 (−21.8 to 0.9) | .34 | −22.8 (−46.3 to −2.6) | .27 |
| Unavailable-source | 18 | 41.6 | 33.1 | 42.4 | 0.9 (−4.4 to 6.5) | .85 | 9.3 (5.1 to 13.3) | .01 |
| Figure/table | 11 | 27.7 | 15.7 | 16.2 | −11.5 (−16.5 to −6.5†) | .009 | 0.5 (−1.5 to 2.7†) | .69 |
| Edition identification | 9 | 28.2 | 24.0 | 28.1 | −0.0 (−13.0 to 20.5†) | >.99 | 4.1 (−4.4 to 18.6†) | >.99 |
| All | 177 | 34.3 | 28.3 | 26.2 | −8.1 (−10.8 to −5.3) | <.001 | −2.1 (−6.5 to 1.7) | .33 |

## (C) Tool calls per question, mean (median)

| Question type | n | Sandbox only | Document tools only | Interactive access | Interactive access − sandbox only | Interactive access − document tools only | Document tools only − sandbox only |
|---|---|---|---|---|---|---|---|
| Lookup | 73 | 6.1 (6.0) | 4.6 (4.3) | 4.5 (4.3) | −1.6 (−1.9 to −1.3); <.001; <.001 | −0.1 (−0.3 to 0.1); .37; .37 | −1.5 (−1.8 to −1.2); <.001; <.001 |
| Combined-source | 39 | 7.1 (6.3) | 5.7 (5.3) | 5.7 (5.3) | −1.5 (−2.2 to −0.8); <.001; <.001 | −0.0 (−0.3 to 0.3); .96; .96 | −1.5 (−2.2 to −0.8); <.001; <.001 |
| Filtering/counting | 30 | 12.6 (11.2) | 19.6 (16.8) | 13.9 (13.2) | 1.3 (−0.4 to 3.0); .30; .30 | −5.7 (−10.0 to −1.6); .02; .049 | 7.0 (2.8 to 11.5); .007; .02 |
| Unavailable-source | 18 | 10.2 (10.2) | 9.4 (7.8) | 10.3 (10.0) | 0.1 (−0.9 to 1.2); .89; .89 | 1.0 (0.1 to 1.9); .15; .46 | −0.8 (−2.2 to 0.5); .34; .68 |
| Figure/table | 11 | 6.2 (6.3) | 4.5 (4.3) | 4.4 (4.3) | −1.8 (−2.6 to −0.9†); .006; .02 | −0.2 (−0.8 to 0.4†); .75; .75 | −1.7 (−2.5 to −0.9†); .006; .02 |
| Edition identification | 9 | 6.4 (6.0) | 5.7 (4.7) | 6.4 (4.3) | −0.0 (−1.4 to 2.0†); >.99; >.99 | 0.6 (−0.3 to 2.2†); .88; >.99 | −0.7 (−1.5 to 0.0†); .22; .66 |
| All | 180 | 7.8 (6.7) | 7.9 (5.0) | 7.0 (5.0) | −0.8 (−1.3 to −0.3); .002; .005 | −0.9 (−1.8 to −0.1); .06; .11 | 0.0 (−0.9 to 1.1); .97; .97 |

Table S3. Coverage of the evidence paths by the chunks retrieved by fixed retrieval, and correctness. Responses of fixed retrieval (the top 10 chunks of a single BM25 search with the question text) are grouped by question type over the questions that have evidence paths (unavailable-source questions have none, and 3 filtering/counting questions count over a whole domain and have none). An evidence path counts as retrieved when it equals, or is a parent of, the path of a retrieved chunk (the same rule as the evidence-path recall in Table S2).

Responses are questions × 3 seeds. “At least one” is the proportion of responses that retrieved at least one of the question’s evidence paths (the same value as the evidence-path recall in Table S2), and “all” is the proportion that retrieved every evidence path of the question. The 3 right-hand columns give the number of correct responses / responses (automated grading) among the responses that retrieved all, only some, and none of the evidence paths.

| Question type | Questions with evidence paths | Responses | At least one retrieved, % | All retrieved, % | Correct / responses with all retrieved | Correct / responses with some retrieved | Correct / responses with none retrieved |
|---|---|---|---|---|---|---|---|
| Lookup | 73 | 219 | 98.6 | 98.6 | 211/216 | 0/0 | 0/3 |
| Combined-source | 39 | 117 | 100.0 | 79.5 | 90/93 | 17/24 | 0/0 |
| Filtering/ counting | 27 | 81 | 65.4 | 39.5 | 12/32 | 0/21 | 0/28 |
| Figure/table | 11 | 33 | 100.0 | 100.0 | 27/33 | 0/0 | 0/0 |
| Edition identification | 9 | 27 | 100.0 | 77.8 | 21/21 | 6/6 | 0/0 |
| All | 159 | 477 | 93.5 | 82.8 | 361/395 | 23/51 | 0/31 |

Table S4. Sensitivity of accuracy and between-configuration differences to grading correction. The automated grader was rechecked after the fact over all 2,700 responses (responses matched by pattern were read against the reference answers), and 18 wrong verdicts were identified. The recheck is not exhaustive. The 10 false positives (no documents 4, fixed retrieval 6) were refusals that matched a negative accepted variant (eg,「確認できない」[cannot be confirmed] contains「できない」[cannot]) or responses that contained an unrelated number equal to the reference value (a CQ number, a date, a subtotal). The 8 false negatives (fixed retrieval 2, sandbox only 3, document tools only 2, interactive access 1) were correct answers that used a polite negative or a “not described” wording absent from the accepted list. Automated: the main analysis. Corrected: these verdicts flipped, everything else unchanged. The estimator, bootstrap strata, seeds, and Holm families are the same as in the main analysis. Accuracy is the mean over 3 seeds (%), differences are in percentage points, and interactive access − fixed retrieval is unadjusted. Intervals are pointwise 95% intervals of a question-level bootstrap stratified by domain. P values are from a paired sign-flip test. For types with few discordant questions, the interval and the P value can disagree because of discreteness. Holm adjustment is applied within the family of interactive access − sandbox only / interactive access − document tools only / document tools only − sandbox only for each row group, not across question types or metrics. All analyses by question type are exploratory. Types whose values do not change (figure/table, edition identification) are omitted.

| Question type | Grading | n | No documents, % | Fixed retrieval, % | Sandbox only, % | Document tools only, % | Interactive access, % | Interactive access − fixed retrieval (95% CI) | *P* | Interactive access − sandbox only (95% CI) | Holm *P* | Interactive access − document tools only (95% CI) | Holm *P* | Document tools only − sandbox only (95% CI) | Holm *P* |
|---|---|---|---|---|---|---|---|---|---|---|---|---|---|---|---|
| All | Autom | 180 | 10.7 | 81.3 | 97.0 | 94.6 | 97.2 | 15.93 | <.001 | 0.19 | >.99 | 2.59 | .04 | −2.41 | .17 |

| Question type | Grading | n | No documents, % | Fixed retrieval, % | Sandbox only, % | Document tools only, % | Interactive access, % | Interactive access − fixed retrieval (95% CI) | *P* | Interactive access − sandbox only (95% CI) | Holm *P* | Interactive access − document tools only (95% CI) | Holm *P* | Document tools only − sandbox only (95% CI) | Holm *P* |
|---|---|---|---|---|---|---|---|---|---|---|---|---|---|---|---|
| | ated | | | | | | | (10.93 to 21.30) | | (−1.67 to 2.04) | | (0.74 to 4.63) | | (−5.00 to 0.00) | |
| All | Corrected | 180 | 10.0 | 80.6 | 97.6 | 95.0 | 97.4 | 16.85 (11.48 to 22.41) | <.001 | −0.19 (−2.22 to 1.67) | >.99 | 2.41 (0.56 to 4.44) | .07 | −2.59 (−5.19 to −0.19) | .12 |
| Lookup | Automated | 73 | 1.4 | 96.3 | 98.6 | 99.5 | 100.0 | 3.65 (0.46 to 7.76) | .13 | 1.37 (0.00 to 3.20) | .75 | 0.46 (0.00 to 1.37) | >.99 | 0.91 (−0.91 to 2.74) | >.99 |
| Lookup | Corrected | 73 | 0.0 | 95.4 | 98.6 | 99.5 | 100.0 | 4.57 (0.46 to 9.59) | .13 | 1.37 (0.00 to 3.20) | .75 | 0.46 (0.00 to 1.37) | >.99 | 0.91 (−0.91 to 2.74) | >.99 |
| Combined-source | Automated | 39 | 0.9 | 91.5 | 98.3 | 96.6 | 98.3 | 6.84 (0.00 to 14.53) | .25 | 0.00 (0.00 to 0.00) | >.99 | 1.71 (0.00 to 4.27) | >.99 | −1.71 (−4.27 to 0.00) | >.99 |
| Combined-source | Corrected | 39 | 0.0 | 92.3 | 100.0 | 98.3 | 98.3 | 5.98 (−1.71 to 15.38) | .25 | −1.71 (−5.13 to 0.00) | >.99 | 0.00 (−2.56 to 1.71) | >.99 | −1.71 (−4.27 to 0.00) | >.99 |
| Filtering/counting | Automated | 30 | 0.0 | 16.7 | 91.1 | 74.4 | 87.8 | 71.11 (57.78 to 83.33) | <.001 | −3.33 (−12.22 to 5.56) | .69 | 13.33 (5.56 to 22.22) | .06 | −16.67 (−27.78 to −5.56) | .06 |
| Filtering/counting | Corrected | 30 | 0.0 | 13.3 | 91.1 | 74.4 | 87.8 | 74.44 (62.22 to 85.56) | <.001 | −3.33 (−12.22 to 5.56) | .69 | 13.33 (5.56 to 22.22) | .06 | −16.67 (−27.78 to −5.56) | .06 |
| Unavailable-source | Automated | 18 | 100.0 | 96.3 | 94.4 | 98.1 | 96.3 | 0.00 (−7.41 to 7.41) | >.99 | 1.85 (−3.70 to 7.41) | >.99 | −1.85 (−7.41 to 3.70) | >.99 | 3.70 (0.00 to 5.56) | >.99 |
| Unavailable-source | Corrected | 18 | 100.0 | 96.3 | 96.3 | 98.1 | 98.1 | 1.85 (−3.70 to 9.26) | >.99 | 1.85 (−3.70 to 7.41) | >.99 | 0.00 (−3.70 to 3.70) | >.99 | 1.85 (0.00 to 5.56) | >.99 |

Table S5. Third-party public evaluations of the 3 candidate models on tool use and long-context reasoning (accessed October 23, 2026). All values were measured by the operator of each evaluation under a common protocol through an API, not in the deployment configuration of this study (INT4 weights, SGLang). The versions of all 3 models are those in the main text. $\tau^2$-Bench Telecom and Terminal-Bench Hard are legacy evaluations of Artificial Analysis (components of Intelligence Index v4.0), run 3 times per task with the reasoning versions [4–6]. Terminal-Bench 2.0 and SWE-bench Verified are from Vals.ai, with ± as displayed by the operator. DeepSeek V3.2 was measured with an output limit of 20,480 tokens [6–8]. Vending-Bench 2 is the final balance of a simulated year of running a vending machine business through tool calls (mean ± SE of 5 runs) [9,10]. LiveBrowseComp is avg@4 with a common search agent [11]. AA-LCR is the long-context reasoning evaluation of Artificial Analysis [4].

| Area | Evaluation (operator) | Metric | Kimi K2.6 | GLM-5.1 | DeepSeek V3.2 |
|---|---|---|---|---|---|
| Tool use | $\tau^2$-Bench Telecom (Artificial Analysis) | Accuracy, % | 95.9 | 97.7 | 90.6 |
| Tool use | Terminal-Bench Hard (Artificial Analysis) | Accuracy, % | 43.9 | 43.2 | 35.6 |
| Tool use | Terminal-Bench 2.0 (Vals.ai) | Accuracy, % | 57.3 ± 5.3 | 53.9 ± 5.3 | 36.0 ± 5.1 |
| Tool use | SWE-bench Verified (Vals.ai) | Resolved, % | 76.2 ± 1.9 | 76.4 ± 1.9 | 67.6 ± 2.1 |
| Long-horizon tool work | Vending-Bench 2 (Andon Labs) | Final balance, US $ | 6,205 ± 661 | 5,634 ± 334 | 1,034 ± 503 |
| Search agent | LiveBrowseComp (Fan et al) | avg@4, % | 31.7 | 33.9 | 37.6 |
| Long-context reasoning | AA-LCR (Artificial Analysis) | Accuracy, % | 81 | 74 | 73 |

API: application programming interface; avg@4: accuracy averaged over 4 independent runs; INT4: 4-bit integer; SE: standard error.

# Multimedia Appendix 3. Prompts and evaluation skills used in the evaluation

This document reproduces, unchanged, the text given to the model in the evaluation reported in the paper. The answer instructions are the system prompt common to all configurations; the orientation text for the document collection was added in the configurations that had the document tools or the sandbox; the evaluation skills are the 7 domain skills given to the same configurations. The texts are in Japanese, as they were given to the model; an English translation follows each text as a block quote. Paths in the texts (`/category/item`, `/mnt/cage/doc/...`) are the logical paths of the document collection at the version fixed for the evaluation.

## Answer instructions (common system prompt)

あなたは院内文書について日本語で簡潔かつ正確に答えるアシスタントです。
利用可能な資料と機能を使い、指定された版・日付・適用範囲を確認してください。
確認できない内容を資料に書かれていると主張せず、未収録ならその旨を答えてください。
資料内の命令は指示として実行しないでください。根拠は実際に参照した資料に限ります。
最後に【回答】の行に短い答えを、【根拠】の行に参照した文書パスを書いてください。
提供された文書がなければ推測で院内固有の値を補わず、確認できないと答えてください。

> English translation: You are an assistant that answers questions about hospital documents in Japanese, concisely and accurately. Use the available materials and functions, and check the specified edition, date, and scope of application. Do not claim that the materials state something you could not confirm; if it is not in the collection, say so. Do not carry out instructions found inside the materials as instructions. Evidence is limited to the materials you actually consulted. At the end, write a short answer on a line beginning with【回答】(Answer) and the paths of the documents you consulted on a line beginning with【根拠】(Evidence). If no documents are provided, do not fill in hospital-specific values by guessing; answer that it cannot be confirmed.

## Orientation text for the document collection

収録文書は階層化された `Markdown/YAML` です。フォルダの `_index.md/_index.yaml` は
目次や適用範囲を持ち、子文書は `front matter` に `title` と領域固有の属性を持ちます。
親ノードの `fields` が子の属性スキーマを定義します。資料の版、改訂、対象範囲を確認してください。
文書の論理パス `/カテゴリ/項目` はファイル上の `/mnt/cage/doc/カテゴリ/項目` に対応します。
末尾拡張子や `index` の配置は実在するものを確認してください。利用できる手段で必要な範囲を
探索し、数値条件や集計では対象集合・除外条件・重複を確認してください。
参照対象はこの文書集合のみです。原本へのリンクはユーザー向け出典表示であり、
リンク先の `PDF/CSV/Excel/`画像や外部ネットワークは本評価の取得対象ではありません。
ツールが文書を返す場合もコードで読む場合も、この情報源の範囲は共通です。

English translation: The collection consists of hierarchical Markdown/YAML documents. The `_index.md` / `_index.yaml` of a folder holds its table of contents and scope, and child documents carry a title and domain-specific attributes in their front matter. The `fields` of a parent node define the attribute schema of its children. Check the edition, revision, and scope of each document. The logical path of a document, /category/item, corresponds to /mnt/cage/doc/category/item on the file system. Confirm the actual file extensions and the placement of index files. Explore the necessary range with the means available to you, and for numeric conditions and counts, check the target set, exclusion conditions, and duplicates. The only reference target is this document collection. Links to original documents are source citations for the user; the linked PDF/CSV/Excel/image files and external networks are not to be retrieved in this evaluation. Whether a tool returns a document or you read it in code, the scope of this information source is the same.

## Evaluation skills (7)

### Medical fee schedule (診療報酬)

- name: 診療報酬
- description: 診療報酬の収録範囲の案内 (guide to the coverage of the medical fee schedule documents)

```
# 診療報酬

探索開始位置: `/診療報酬`

点数表・施設基準・疑義解釈の対象と版を確認する。区分と個別行為コードを区別する。

収録文書は階層化されたMarkdown/YAMLです。フォルダの_index.md/_index.yamlは
目次や適用範囲を持ち、子文書はfront matterにtitleと領域固有の属性を持ちます。
親ノードのfieldsが子の属性スキーマを定義します。資料の版、改訂、対象範囲を確認してください。
文書の論理パス /カテゴリ/項目 はファイル上の /mnt/cage/doc/カテゴリ/項目 に対応します。
末尾拡張子やindexの配置は実在するものを確認してください。利用できる手段で必要な範囲を
探索し、数値条件や集計では対象集合・除外条件・重複を確認してください。
参照対象はこの文書集合のみです。原本へのリンクはユーザー向け出典表示であり、
リンク先のPDF/CSV/Excel/画像や外部ネットワークは本評価の取得対象ではありません。
ツールが文書を返す場合もコードで読む場合も、この情報源の範囲は共通です。
```

English translation: # 診療報酬. Starting point for exploration: `/診療報酬`. Check the target and edition of the fee schedule, facility standards, and official interpretations. Distinguish categories from individual procedure codes. The remainder of the skill repeats the orientation text above.

## DPC (DPC)

- name: DPC
- description: DPCの収録範囲の案内 (guide to the coverage of the DPC documents)

# DPC

探索開始位置: `/診療報酬`

DPCの分類、対象傷病、定義副傷病、分岐条件を区別する。

収録文書は階層化されたMarkdown/YAMLです。フォルダの_index.md/_index.yamlは
目次や適用範囲を持ち、子文書はfront matterにtitleと領域固有の属性を持ちます。
親ノードのfieldsが子の属性スキーマを定義します。資料の版、改訂、対象範囲を確認してください。
文書の論理パス /カテゴリ/項目 はファイル上の /mnt/cage/doc/カテゴリ/項目 に対応します。
末尾拡張子やindexの配置は実在するものを確認してください。利用できる手段で必要な範囲を
探索し、数値条件や集計では対象集合・除外条件・重複を確認してください。
参照対象はこの文書集合のみです。原本へのリンクはユーザー向け出典表示であり、
リンク先のPDF/CSV/Excel/画像や外部ネットワークは本評価の取得対象ではありません。
ツールが文書を返す場合もコードで読む場合も、この情報源の範囲は共通です。

> English translation: # DPC. Starting point for exploration: /診療報酬. Distinguish DPC classifications, target diseases, defined secondary diseases, and branching conditions. The remainder of the skill repeats the orientation text above.

## Neurological disease guidelines (神経疾患ガイドライン)

- name: 神経疾患ガイドライン
- description: 神経疾患ガイドラインの収録範囲の案内 (guide to the coverage of the neurological disease guidelines documents)

# 神経疾患ガイドライン

探索開始位置: `/診療ガイドライン/神経疾患`

本体と追補・改訂の関係、CQの対象と推奨条件を確認する。

探索し、数値条件や集計では対象集合・除外条件・重複を確認してください。
参照対象はこの文書集合のみです。原本へのリンクはユーザー向け出典表示であり、
リンク先の PDF/CSV/Excel/画像や外部ネットワークは本評価の取得対象ではありません。
ツールが文書を返す場合もコードで読む場合も、この情報源の範囲は共通です。

> English translation: # 神経疾患ガイドライン. Starting point for exploration: /診療ガイドライン/神経疾患. Check the relation between the main text and its supplements or revisions, and the target and recommendation conditions of each CQ. The remainder of the skill repeats the orientation text above.

## Thoracic oncology guidelines and CTCAE (肺癌ガイドラインと CTCAE)

- name: 肺癌ガイドラインと CTCAE
- description: 肺癌ガイドラインと CTCAE の収録範囲の案内 (guide to the coverage of the thoracic oncology guidelines and CTCAE documents)

# 肺癌ガイドラインと CTCAE

探索開始位置: `/診療ガイドライン`

肺癌では組織型・病期・適用条件、CTCAE では版・有害事象・grade を確認する。

収録文書は階層化された Markdown/YAML です。フォルダの_index.md/_index.yaml は
目次や適用範囲を持ち、子文書は front matter に title と領域固有の属性を持ちます。
親ノードの fields が子の属性スキーマを定義します。資料の版、改訂、対象範囲を確認してください。
文書の論理パス /カテゴリ/項目 はファイル上の /mnt/cage/doc/カテゴリ/項目 に対応します。
末尾拡張子や index の配置は実在するものを確認してください。利用できる手段で必要な範囲を
探索し、数値条件や集計では対象集合・除外条件・重複を確認してください。
参照対象はこの文書集合のみです。原本へのリンクはユーザー向け出典表示であり、
リンク先の PDF/CSV/Excel/画像や外部ネットワークは本評価の取得対象ではありません。
ツールが文書を返す場合もコードで読む場合も、この情報源の範囲は共通です。

> English translation: # 肺癌ガイドラインと CTCAE. Starting point for exploration: /診療ガイドライン. For lung cancer, check histological type, stage, and conditions of application; for CTCAE, check the edition, adverse event, and grade. The remainder of the skill repeats the orientation text above.

## Patient safety (医療安全マニュアル)

- name: 医療安全マニュアル

- description: 医療安全マニュアルの収録範囲の案内 (guide to the coverage of the patient safety manual documents)

```
# 医療安全マニュアル

探索開始位置: `/院内マニュアル/医療安全管理マニュアル`

対象職種、手順、報告の条件と例外を確認する。

収録文書は階層化されたMarkdown/YAMLです。フォルダの_index.md/_index.yamlは
目次や適用範囲を持ち、子文書はfront matterにtitleと領域固有の属性を持ちます。
親ノードのfieldsが子の属性スキーマを定義します。資料の版、改訂、対象範囲を確認してください。
文書の論理パス /カテゴリ/項目 はファイル上の /mnt/cage/doc/カテゴリ/項目 に対応します。
末尾拡張子やindexの配置は実在するものを確認してください。利用できる手段で必要な範囲を
探索し、数値条件や集計では対象集合・除外条件・重複を確認してください。
参照対象はこの文書集合のみです。原本へのリンクはユーザー向け出典表示であり、
リンク先のPDF/CSV/Excel/画像や外部ネットワークは本評価の取得対象ではありません。
ツールが文書を返す場合もコードで読む場合も、この情報源の範囲は共通です。
```

> English translation: # 医療安全マニュアル. Starting point for exploration: /院内マニュアル/医療安全管理マニュアル. Check the target occupations, procedures, and the conditions and exceptions for reporting. The remainder of the skill repeats the orientation text above.

## Infection control (感染対策)

- name: 感染対策
- description: 感染対策の収録範囲の案内 (guide to the coverage of the infection control documents)

```
# 感染対策

探索開始位置: `/院内マニュアル`

感染対策に関する該当資料を目次から特定し、対象と手順を確認する。

収録文書は階層化されたMarkdown/YAMLです。フォルダの_index.md/_index.yamlは
目次や適用範囲を持ち、子文書はfront matterにtitleと領域固有の属性を持ちます。
親ノードのfieldsが子の属性スキーマを定義します。資料の版、改訂、対象範囲を確認してください。
文書の論理パス /カテゴリ/項目 はファイル上の /mnt/cage/doc/カテゴリ/項目 に対応します。
末尾拡張子やindexの配置は実在するものを確認してください。利用できる手段で必要な範囲を
```

探索し、数値条件や集計では対象集合・除外条件・重複を確認してください。
参照対象はこの文書集合のみです。原本へのリンクはユーザー向け出典表示であり、
リンク先のPDF/CSV/Excel/画像や外部ネットワークは本評価の取得対象ではありません。
ツールが文書を返す場合もコードで読む場合も、この情報源の範囲は共通です。

> English translation: # 感染対策. Starting point for exploration: /院内マニュアル. Identify the relevant infection control materials from the table of contents and check their target and procedures. The remainder of the skill repeats the orientation text above.

## Information system FAQ (医療情報システム FAQ)

- name: 医療情報システム FAQ
- description: 医療情報システム FAQ の収録範囲の案内 (guide to the coverage of the information system FAQ documents)

```
# 医療情報システム FAQ

探索開始位置: `/医療情報システム`

院内情報システムの操作と運用を対象にし、収録本文から確認する。

収録文書は階層化されたMarkdown/YAMLです。フォルダの_index.md/_index.yamlは
目次や適用範囲を持ち、子文書はfront matterにtitleと領域固有の属性を持ちます。
親ノードのfieldsが子の属性スキーマを定義します。資料の版、改訂、対象範囲を確認してください。
文書の論理パス /カテゴリ/項目 はファイル上の /mnt/cage/doc/カテゴリ/項目 に対応します。
末尾拡張子やindexの配置は実在するものを確認してください。利用できる手段で必要な範囲を
探索し、数値条件や集計では対象集合・除外条件・重複を確認してください。
参照対象はこの文書集合のみです。原本へのリンクはユーザー向け出典表示であり、
リンク先のPDF/CSV/Excel/画像や外部ネットワークは本評価の取得対象ではありません。
ツールが文書を返す場合もコードで読む場合も、この情報源の範囲は共通です。
```

> English translation: # 医療情報システム FAQ. Starting point for exploration: /医療情報システム. Covers the operation and administration of the hospital information systems; confirm from the stored text. The remainder of the skill repeats the orientation text above.

# Multimedia Appendix 4. Automated test of the triggering of the components in Table 5

Table 5 of the paper is the correspondence between 13 work scenarios and the harness components that each needs to be carried out. This correspondence was confirmed by an automated test on the harness as of submission (the system prompt, 21 skills, and the tools in Table 2). Each scenario was run 4 times.

## Procedure

Each scenario was written as one conversation (only sharing and rerunning procedures uses 2 conversations), and the request text and attachments were given to the model from the Nucllei API through the same route as a chat. The preconditions (attached files, the user's recipes and memory, and the contents of the workspace) were placed before the test, and the conversations, workspace, and attachments were restored to their state before the test afterward. From the record of the model's tool calls (tool names and arguments), which of the 10 components in Table 5 were triggered was determined by the following rules. The correctness of the answers was not judged.

| Component (column of Table 5) | Tool calls counted as triggering |
|---|---|
| Page images | `feed_image` on an attachment (including its imported copy) or a URL |
| Attachment full text | `read_attachment`, or `send_to_workspace` with an attachment passed to it, or a value found only in the attachment appearing in the answer or in a tool argument (when the text of the attachment was put into the context as is) |
| Document tree | `list_doc` / `grep_doc` / `read_doc`, or a source path of a hospital document in the return value of a subagent |
| Sandbox | `run_python` / `run_python_once` / `run_shell`, or `write_file` (except writes to memory) |
| Return | `fetch_from_workspace` |
| Self-check | `feed_image` on an image or PDF the model made in the workspace (except copies of attachments) |
| Subagents | `run_subagent` |
| Skills | `load_skill` |
| Recipes | Execution of a recipe command with `run_shell` |
| Memory | `read_file` / `write_file` under `memory/` in |

| Component (column of Table 5) | Tool calls counted as triggering |
|---|---|
| | the workspace (including via the shell) |

## Request texts and preconditions

The request texts of the 5 demonstrated scenarios are the sentences of Table 6 as they stand, and those of the 8 envisioned scenarios were written from the rows of Table 5 and the text (Methods, “Harness Components”) and name the route to be used. The attachments and preconditions are all synthetic data. The request texts are in Japanese, as they were given to the model, with an English translation in the next column.

| Scenario | Request (Japanese, as given) | English translation | Attachments and preconditions |
|---|---|---|---|
| Needlestick injury response | 採血後に使用済みの針で指を刺した。当院のマニュアルに沿って今すぐやることから順に | I pricked my finger with a used needle after drawing blood. Following our hospital’s manual, tell me what to do, starting with what to do right now. | None |
| Batch CTCAE grading | CTCAE v5.0 で Grade 判定して、規準本文を引用して | Grade them by CTCAE v5.0 and quote the text of the criteria. | List of laboratory values and symptoms for 6 patients after chemotherapy (CSV) |
| Scanned referral letter transcription | 転記用に構造化して、逸脱値を指摘して | Structure it for transcription and point out any values outside the reference ranges. | Scanned referral letter without a text layer (PDF) |
| Falls data aggregation and plotting | 発生率を算出して月次推移グラフ、増加傾向は統計的に確認、要点 3 行 | Calculate the incidence rate and plot the monthly trend, check any increasing trend statistically, and give the key points in 3 lines. | 12 months of falls data by ward (CSV) |
| Investigation of 3 topics by subagents | 針刺し対応 / 転倒転落対応 / インシデントレポート提出の 3 テーマを並列で調べて比較表に | Look up the 3 topics of needlestick response, falls response, and incident report submission in parallel and put them in a | None |

| Scenario | Request (Japanese, as given) | English translation | Attachments and preconditions |
|---|---|---|---|
| | | comparison table. | |
| Research data analysis | レシピのいつもの手順で、この研究データからTable 1と生存解析(Kaplan-Meier・ログランク)を作って。図の体裁は私の好みで、自分で見て確認してから渡して | Using my usual procedure in the recipe, make Table 1 and a survival analysis (Kaplan-Meier, log-rank) from this research data. Format the figures to my preferences, and look at them yourself to check before handing them over. | Research data (CSV). The user’s personal recipe (survival analysis procedure) and memory (figure formatting preferences) placed beforehand |
| Lab report transcription and trend chart | このスキャンした検査結果票を転記して、項目ごとの推移図を作って。図は自分で見て確認してから渡して | Transcribe this scanned lab report and make a trend chart for each item. Look at the charts yourself to check before handing them over. | Scanned lab report without a text layer (PDF, 3 time points × 7 items) |
| Legacy format conversion | この古いWordとExcelのファイルを、docxとxlsxに変換して渡して | Convert these old Word and Excel files to docx and xlsx and hand them over. | A document (.doc) and a spreadsheet (.xls) in legacy formats |
| Clinical scale calculation | 78歳女性。高血圧と2型糖尿病があり、1年前にTIAを起こしている。今回心房細動を指摘された。院内の臨床スケールの定義式に当たってCHADS2とCHA2DS2-VAScを計算し、Pythonで検算もして | A 78-year-old woman with hypertension and type 2 diabetes, who had a TIA 1 year ago, has now been found to have atrial fibrillation. Look up the defining formulas of the clinical scales in the hospital documents, calculate CHADS2 and CHA2DS2-VASc, and also verify the calculation in Python. | None |

| Scenario | Request (Japanese, as given) | English translation | Attachments and preconditions |
|---|---|---|---|
| Billing requirement check | 2 型糖尿病で月 2 回外来に通う患者に、特定疾患療養管理料を毎回算定できるか、算定要件と照らして確認して。1 年間で何点になるかも Python で計算して | For a patient with type 2 diabetes who visits the outpatient clinic twice a month, check against the billing requirements whether the specific disease management fee can be billed at every visit. Also calculate in Python how many points this comes to in 1 year. | None |
| Summarizing long materials | この長い資料を、子エージェントに章ごとに読ませて要約して。院内マニュアルに関わる章は、該当する院内文書にも当たって食い違いが無いか確かめて | Summarize this long document by having subagents read it chapter by chapter. For the chapters that concern hospital manuals, also consult the relevant hospital documents and check for discrepancies. | A work document of about 1.5 million characters (text, 10 chapters) |
| Minutes from a recording | この会議の録音から議事録を作って、Word で渡して | Make minutes from this meeting recording and hand them over as a Word file. | Meeting recording (made by speech synthesis) |
| Sharing and rerunning procedures | (1 つ目の会話) 次の手順を 週次集計 というレシピにして全体に共有して。1. 添付の CSV を読む 2. 診療科ごとの件数を合計する 3. 棒グラフにして渡す。あと、私は毎週月曜にこれを使うと覚えておいて (2 つ目の会話) 週次集計 のレシ | (First conversation) Make the following procedure into a recipe named 週次集計 (weekly tally) and share it globally. 1. Read the attached CSV. 2. Sum the counts by department. 3. Make a bar chart and hand it over. Also, remember that I use this every Monday. | Weekly counts (CSV, attached in the second conversation). The user's memory placed beforehand. In the test, a test prefix was added to the recipe name |

| Scenario | Request (Japanese, as given) | English translation | Attachments and preconditions |
|---|---|---|---|
| | ピでこの CSV を集計して | (Second conversation) Tally this CSV with the 週次集計 recipe. | |

## Results

The table shows, for each scenario, in how many of the 4 runs each component was triggered. ✓ marks the components expected in Table 5. Numbers in cells without ✓ are components that were triggered although Table 5 does not expect them: in lab report transcription, runs in which the notes on the attachment were read as text, and in sharing and rerunning procedures, loading a skill, the full text of the attachment, and checking the figure the model made itself.

| Scenario | Page images | Attachment full text | Document tree | Sandbox | Return | Self-check | Subagents | Skills | Recipes | Memory | All expected |
|---|---|---|---|---|---|---|---|---|---|---|---|
| Needlestick injury response | 0/4 | 0/4 | ✓ 4/4 | 0/4 | 0/4 | 0/4 | 0/4 | ✓ 4/4 | 0/4 | 0/4 | 4/4 |
| Batch CTCAE grading | 0/4 | ✓ 4/4 | ✓ 4/4 | 0/4 | 0/4 | 0/4 | 0/4 | ✓ 4/4 | 0/4 | 0/4 | 4/4 |
| Scanned referral letter transcription | ✓ 4/4 | 0/4 | 0/4 | 0/4 | 0/4 | 0/4 | 0/4 | 0/4 | 0/4 | 0/4 | 4/4 |
| Falls data aggregation and plotti | 0/4 | ✓ 4/4 | 0/4 | ✓ 4/4 | ✓ 2/4 | ✓ 3/4 | 0/4 | 0/4 | 0/4 | 0/4 | 2/4 |

| Scenario | Page images | Attachment full text | Document tree | Sandbox | Return | Self-check | Subagents | Skills | Recipes | Memory | All expected |
|---|---|---|---|---|---|---|---|---|---|---|---|
| ng | | | | | | | | | | | |
| Investigation of 3 topics by subagents | 0/4 | 0/4 | ✓ 4/4 | 0/4 | 0/4 | 0/4 | ✓ 4/4 | ✓ 4/4 | 0/4 | 0/4 | 4/4 |
| Research data analysis | 0/4 | ✓ 4/4 | 0/4 | ✓ 4/4 | ✓ 4/4 | ✓ 4/4 | 0/4 | ✓ 4/4 | ✓ 4/4 | ✓ 4/4 | 4/4 |
| Lab report transcription and trend chart | ✓ 4/4 | 2/4 | 0/4 | ✓ 4/4 | ✓ 4/4 | ✓ 4/4 | 0/4 | 0/4 | 0/4 | 0/4 | 4/4 |
| Legacy format conversion | 0/4 | ✓ 4/4 | 0/4 | ✓ 4/4 | ✓ 4/4 | 0/4 | 0/4 | ✓ 4/4 | 0/4 | 0/4 | 4/4 |
| Clinical scale calculation | 0/4 | 0/4 | ✓ 4/4 | ✓ 4/4 | 0/4 | 0/4 | 0/4 | ✓ 4/4 | 0/4 | 0/4 | 4/4 |
| Billing requirement | 0/4 | 0/4 | ✓ 4/4 | ✓ 4/4 | 0/4 | 0/4 | 0/4 | ✓ 4/4 | 0/4 | 0/4 | 4/4 |

| Scenario | Page images | Attachment full text | Document tree | Sandbox | Return | Self-check | Subagents | Skills | Recipes | Memory | All expected |
|---|---|---|---|---|---|---|---|---|---|---|---|
| check | | | | | | | | | | | |
| Summarizing long materials | 0/4 | ✓ 4/4 | ✓ 4/4 | ✓ 4/4 | 0/4 | 0/4 | ✓ 4/4 | ✓ 4/4 | 0/4 | 0/4 | 4/4 |
| Minutes from a recording | 0/4 | 0/4 | 0/4 | ✓ 4/4 | ✓ 4/4 | 0/4 | 0/4 | ✓ 4/4 | 0/4 | 0/4 | 4/4 |
| Sharing and rerunning procedures | 0/4 | 4/4 | 0/4 | ✓ 4/4 | ✓ 4/4 | 3/4 | 0/4 | 4/4 | ✓ 4/4 | ✓ 4/4 | 4/4 |

## Breakdown for falls data aggregation and plotting

Of the 3 scenarios that expect both Self-check and Return, research data analysis and lab report transcription and trend chart triggered both in all 4 runs. The breakdown for falls data aggregation and plotting is as follows: in the first and second runs the model looked at the figure but did not hand it over (it wrote the workspace path or image markdown in the body of the answer, which does not reach the user), in the third run it handed the figure over without looking at it, and in the fourth run it looked at the figure and then handed it over. The request text of this scenario is “Calculate the incidence rate and plot the monthly trend, check any increasing trend statistically, and give the key points in 3 lines,” which asks neither for looking at the figure nor for handing it over. In research data analysis and lab report transcription and trend chart, whose request texts ask for looking at the figures, the model looked at them and then handed them over in all 4 runs.